\documentclass[runningheads]{llncs}
\usepackage[width=122mm,left=12mm,paperwidth=146mm,height=193mm,top=12mm,paperheight=217mm]{geometry}

\usepackage{graphicx,amssymb,bm,hyperref,multirow}
\usepackage{amsmath,amssymb} % define this before the line numbering.
\usepackage{color}

\newcommand{\etal}{et al.~}
\begin{document}
\pagestyle{headings}
\mainmatter

%===========================================================
\def\ACCV16SubNumber{14}  % Insert your submission number here

\title{Fitting a 3D Morphable Model to Edges: \\A Comparison Between Hard and Soft Correspondences}

\titlerunning{Fitting a 3D Morphable Model to Edges}
\authorrunning{Anil Bas, William A. P. Smith, Timo Bolkart, Stefanie Wuhrer}

\author{Anil Bas$^{\star}$ \qquad William A. P. Smith$^{\star}$ \qquad Timo Bolkart$^{\dagger}$ \qquad Stefanie Wuhrer$^{\ddagger}$}
\institute{$^{\star}$ Department of Computer Science, University of York, UK \\ {\tt \{ab1792, william.smith\}@york.ac.uk}\\
    $^{\dagger}$ Multimodal Computing and Interaction, Saarland University, Germany \\ {\tt tbolkart@mmci.uni-saarland.de} \\
    $^{\ddagger}$ Morpheo Team, INRIA Grenoble Rh\^{o}ne-Alpes, France \\ {\tt stefanie.wuhrer@inria.fr}
    }

\maketitle
%===========================================================
\begin{abstract}
In this paper we explore the problem of fitting a 3D morphable model to single face images using only sparse geometric features (edges and landmark points). Previous approaches to this problem are based on nonlinear optimisation of an edge-derived cost that can be viewed as forming soft correspondences between model and image edges. We propose a novel approach, that explicitly computes hard correspondences. The resulting objective function is non-convex but we show that a good initialisation can be obtained efficiently using alternating linear least squares in a manner similar to the iterated closest point algorithm. We present experimental results on both synthetic and real images and show that our approach outperforms methods that use soft correspondence and other recent methods that rely solely on geometric features.
\end{abstract}
%===========================================================
\section{Introduction}

Estimating 3D face shape from one or more 2D images is a longstanding problem in computer vision. It has a wide range of applications from pose-invariant face recognition \cite{blanz2003face} to creation of 3D avatars from 2D images \cite{ichim2015dynamic}. One of the most successful approaches to this problem is to use a statistical model of 3D face shape \cite{BlanzVetter1999}. This transforms the problem of shape estimation to one of model fitting and provides a strong statistical prior to constrain the problem.

The model fitting objective can be formulated in various ways, the most obvious being an analysis-by-synthesis approach in which appearance error is directly optimised \cite{BlanzVetter1999}. However, feature-based methods \cite{Romdhani:05,Huber2015} are in general more robust and lead to optimisation problems less prone to convergence on local minima. In this paper, we focus on fitting to edge features in images.

Image edges convey important information about a face. The occluding boundary provides direct information about 3D shape, for example a profile view reveals strong information about the shape of the nose. Internal edges, caused by texture changes, high curvature or self occlusion, provide information about the position and shape of features such as lips, eyebrows and the nose. This information provides a cue for estimating 3D face shape from 2D images or, more generally, for fitting face models to images.

In Section \ref{sec:prelim} we introduce relevant background. In Section \ref{sec:fitting} we present a method for fitting to landmarks with known model correspondence. Our key contribution is in Section \ref{sec:hard} where we present a novel, fully automatic algorithm for fitting to image edges with hard correspondence. By hard correspondence, we mean that an explicit correspondence is computed between projected model vertex and edge pixel. For comparison, in Section \ref{sec:soft} we describe our variant of previous methods \cite{Romdhani:05,Amberg2007,keller20073d} that fit to edges using soft correspondence. By soft correspondence, we mean that an energy term that captures many possible edge correspondences is minimised. Finally, we compare the two approaches experimentally and others from the recent literature in Section \ref{sec:exp}.

\subsection{Related Work}

\noindent {\bf Landmark fitting}\ \ \ 
2D landmarks have long been used as a way to initialize a morphable model fit~\cite{BlanzVetter1999}. Breuer~\etal \cite{breuer2008automatic} obtained this initialisation using a landmark detector providing a fully automatic system. More recently, landmarks have been shown to be sufficient for obtaining useful shape estimates in their own right~\cite{aldrian2013inverse}. Furthermore, noisily detected landmarks can be filtered using a model~\cite{amberg2011optimal} and automatic landmark detection can be integrated into a fitting algorithm~\cite{schonborn2013monte}. In a similar manner to landmarks, local features can be used to aid the fitting process~\cite{Huber2015}. 

\noindent {\bf Edge fitting}\ \ \ 
An early example of using image edges for face model fitting is the Active Shape Model (ASM) \cite{Cootes:95} where a 2D boundary model is aligned to image edges. In 3D, contours have been used directly for 3D face shape estimation \cite{Atkinson:09} and indirectly as a feature for fitting a 3DMM. The earliest work in this direction was due to Moghaddam~\etal \cite{moghaddam2003model} who fitted a 3DMM to silhouettes extracted from multiple views. From a theoretical standpoint, L{\"u}thi~\etal \cite{luthi2009probabilistic} explored to what degree face shape is constrained when contours are fixed.

Romdhani~\etal \cite{Romdhani:05} include an edge distance cost as part of a hybrid energy function. Texture and outer (silhouette) contours are used in a similar way to LM-ICP \cite{Fitzgibbon:03} where correspondence between image edges and model contours is ``soft''. This is achieved by applying a distance transform to an edge image. This provides a smoothly varying cost surface whose value at a pixel indicates the distance (and its gradient, the direction) to the closest edge. This idea was extended by Amberg~\etal \cite{Amberg2007} who use it in a multi-view setting and smooth the edge distance cost by averaging results with different parameters. In this way, the cost surface also encodes the saliency of an edge. Keller~\etal \cite{keller20073d} showed that such approaches lead to a cost function that is neither continuous nor differentiable. This suggests the optimisation method must be carefully chosen. 

Edge features have also been used in other ways. Cashman and Fitzgibbon \cite{Cashman:13} learn a 3DMM from 2D images by fitting to silhouettes. Zhu \etal \cite{Zhu:15} present a method that can be seen as a hybrid of landmark and edge fitting. Landmarks that define boundaries are allowed to slide over the 3D face surface during fitting. A recent alternative to optimisation-based approaches is to learn a regressor  from extracted face contours to 3DMM shape parameters \cite{Dalila:16}. 

Fitting a 3DMM to a 2D image using only geometric features (i.e. landmarks and edges) is essentially a non-rigid alignment problem. Surprisingly, the idea of employing an iterated closest point \cite{Besl:92} approach with hard edge correspondences (in a similar manner to ASM fitting) has been discounted in the literature \cite{Romdhani:05}. In this paper, we pursue this idea and develop an iterative 3DMM fitting algorithm that is fully automatic, simple and efficient (and we make our implementation available\footnote{Matlab implementation: \href{http://github.com/waps101/3DMM_edges}{github.com/waps101/3DMM\_edges}}). Instead of working in a transformed distance-to-edge space and treating correspondences as ``soft'', we compute an explicit correspondence between model and image edges. This allows us to treat the model edge vertices as a landmark with known 2D position, for which optimal pose or shape estimates can be easily computed. 

\noindent {\bf State of the art}\ \ \ 
The most recent face shape estimation methods are able to obtain considerably higher quality results than the purely model-based approaches above. They do so by using pixel-wise shading or motion information to apply finescale refinement to an initial shape estimate. For example, Suwajanakorn \etal \cite{Suwajanakorn2014} use photo collections to build an average model of an individual which is then fitted to a video and finescale detail added by optical flow and shape-from-shading. Cao \etal \cite{Cao:15} take a machine learning approach and train a regressor that predicts high resolution shape detail from local appearance.

Our aim in this paper is not to compete directly with these methods. Rather, we seek to understand what quality of face reconstruction it is possible to obtain using solely sparse, geometric information. The output of our method may provide a better initialisation for state of the art refinement techniques or remove the need to have a person specific model.

\section{Preliminaries}\label{sec:prelim}
Our approach is based on fitting a 3DMM to face images under the assumption of a scaled orthographic projection. Hence, we begin by introducing scaled orthographic projection and 3DMMs.

\subsection{Scaled Orthographic Projection}
The scaled orthographic, or weak perspective, projection model assumes that variation in depth over the object is small relative to the mean distance from camera to object. Under this assumption, the projected 2D position of a 3D point ${\bf v}=[u\ v\ w]^{\textrm{T}}$ given by $\textrm{\bf SOP}[{\bf v},{\bf R},{\bf t},s] \in \mathbb{R}^2$ does not depend on the distance of the point from the camera, but only on a uniform scale $s$ given by the ratio of the focal length of the camera and the mean distance from camera to object:
\begin{small}
\begin{equation}
 \textrm{\bf SOP}[{\bf v},{\bf R},{\bf t},s] = s\left[ 
\begin{array}{ccc}
 1 & 0 & 0 \\
 0 & 1 & 0
\end{array}\right]
  {\bf Rv}+s{\bf t}
\end{equation}
\end{small}
where the pose parameters ${\bf R}\in\mathbb{R}^{3\times 3}$, ${\bf t}\in\mathbb{R}^2$ and $s\in\mathbb{R}^+$ are a rotation matrix, 2D translation and scale respectively.

\subsection{3D Morphable Model}
A 3D morphable model is a deformable mesh whose shape is determined by the shape parameters ${\bm \alpha}\in \mathbb{R}^{S}$. Shape is described by a linear model learnt from data using Principal Components Analysis (PCA). So, the shape of any face can be approximated as:
\begin{equation}
{\bf f}({\bm \alpha})= {\bf P}{\bm \alpha}+\bar{\bf f}, \label{eqn:shapemodel}
\end{equation}
where ${\bf P}\in\mathbb{R}^{3N\times S}$ contains the $S$ principal components, $\bar{\bf f}\in\mathbb{R}^{3N}$ is the mean shape and the vector ${\bf f}({\bm \alpha})\in \mathbb{R}^{3N}$ contains the coordinates of the $N$ vertices, stacked to form a long vector: ${\bf f}=\left[u_{1}\ v_{1}\ w_{1}\ \dots\ u_{N}\ v_{N}\ w_{N}\right]^{\textrm{T}}$. Hence, the $i$th vertex is given by: ${\bf v}_{i}=\left[f_{3i-2}\ f_{3i-1}\ f_{3i}\right]^{\textrm{T}}$. 
For convenience, we denote the sub-matrix corresponding to the $i$th vertex as ${\bf P}_i\in\mathbb{R}^{3\times S}$ and the corresponding vertex in the mean face shape as $\bar{\bf f}_i\in\mathbb{R}^3$, such that the $i$th vertex is given by:
$
{\bf v}_i = {\bf P}_i{\bm \alpha}+\bar{\bf f}_i.
$
Similarly, we define the row corresponding to the $u$ component of the $i$th vertex as ${\bf P}_{iu}$ (similarly for $v$ and $w$) and define the $u$ component of the $i$th mean shape vertex as $\bar{f}_{iu}$ (similarly for $v$ and $w$).

\section{Fitting with Known Correspondence}\label{sec:fitting}

We begin by showing how to fit a morphable model to $L$ observed 2D positions ${\bf x}_i=\left[x_i\ y_i\right]^{\textrm{T}}$ ($i=1\dots L$) arising from the projection of corresponding vertices in the morphable model. We discuss in Section \ref{sec:hard} how these correspondences are obtained in practice. Without loss of generality, we assume that the $i$th 2D position corresponds to the $i$th vertex in the morphable model. The objective of fitting a morphable model to these observations is to obtain the shape and pose parameters that minimise the reprojection error, $E_{\textrm{lmk}}$, between observed and predicted 2D positions:
\begin{equation}
E_{\textrm{lmk}}({\bm \alpha},{\bf R},{\bf t},s)=\frac{1}{L} \sum_{i=1}^L \| {\bf x}_i - \textrm{\bf SOP}\left[{\bf P}_i{\bm \alpha}+\bar{\bf f}_i,{\bf R},{\bf t},s\right] \|^2.\label{eqn:reprojerr}
\end{equation}
The scale factor in front of the summation makes the magnitude of the error invariant to the number of landmarks.
This problem is multilinear in the shape parameters and the SOP transformation matrix. It is also nonlinearly constrained, since ${\bf R}$ must be a valid rotation matrix. Although minimising $E_{\textrm{lmk}}$ is a non-convex optimisation problem, a good initialisation can be obtained using alternating linear least squares and this estimate subsequently refined using nonlinear optimisation. This is the approach that we take.

\subsection{Pose Estimation}
We make an initial estimate of ${\bf R}$, ${\bf t}$ and $s$ using a simple extension of the POS algorithm \cite{dementhon1995model}. Compared to POS, we additionally enforce that ${\bf R}$ is a valid rotation matrix. We begin by solving an unconstrained system in a least squares sense. We stack two copies of the 3D points in homogeneous coordinates, such that ${\bf A}_{2i-1}=\left[ u_i\ v_i\ w_i\ 1\ 0\ 0\ 0\ 0\right]$ and ${\bf A}_{2i}=\left[ 0\ 0\ 0\ 0\ u_i\ v_i\ w_i\ 1\right]$ and form a long vector of the corresponding 2D points ${\bf d}=\left[ x_1\ y_1\ \cdots \ x_L\ y_L \right]^{\textrm{T}}$. We then solve for ${\bf k}\in \mathbb{R}^8$ in ${\bf Ak}={\bf d}$ using linear least squares. We define ${\bf r}_1=\left[ {k}_1\ {k}_2\ {k}_3 \right]$ and ${\bf r}_2=\left[ {k}_5\ {k}_6\ {k}_7 \right]$. Scale is given by $s=(\|{\bf r}_1\|+\|{\bf r}_2\|)/2$ and the translation vector by ${\bf t}=\left[ {k}_4/s\ {k}_8/s \right]^{\textrm{T}}$.
We perform singular value decomposition on the matrix formed from ${\bf r}_1$ and ${\bf r}_2$:
\begin{equation}
{\bf USV}^{\textrm{T}}=\left[ \begin{array}{c}
{\bf r}_1 \\
{\bf r}_2 \\
{\bf r}_1 \times {\bf r}_2
\end{array}
\right]
\end{equation}
The rotation matrix is given by ${\bf R}={\bf UV}^{\textrm{T}}$. If $\textrm{det}({\bf R})=-1$ then we negate the third row of ${\bf U}$ and recompute ${\bf R}$. This guarantees that ${\bf R}$ is a valid rotation matrix.
This approach gives a good initial estimate which we subsequently refine with nonlinear optimization of $E_{\textrm{lmk}}$ with respect to ${\bf R}$, ${\bf t}$ and $s$.

\subsection{Shape Estimation}
With a fixed pose estimate, shape parameter estimation under scaled orthographic projection is a linear problem.  The 2D position of the $i$th vertex as a function of the shape parameters is given by:
$
s{\bf R}_{1..2}({\bf P}_i{\bm \alpha}+\bar{\bf f}_i)+s{\bf t}.
$
Hence, each observed vertex adds two equations to a linear system. Concretely, for each image we form the matrix ${\bf C}\in\mathbb{R}^{2L\times S}$ where
$${\bf C}_{2i-1}=s({\bf R}_{11}{\bf P}_{iu}^{\textrm{T}}+{\bf R}_{12}{\bf P}_{iv}^{\textrm{T}}+{\bf R}_{13}{\bf P}_{iw}^{\textrm{T}})$$ and
$${\bf C}_{2i}=s({\bf R}_{21}{\bf P}_{iu}^{\textrm{T}}+{\bf R}_{22}{\bf P}_{iv}^{\textrm{T}}+{\bf R}_{23}{\bf P}_{iw}^{\textrm{T}})$$
and vector ${\bf h}\in\mathbb{R}^{2L}$ where
$${\bf h}_{2i-1}=x_{i}-s({\bf R}_1\bar{\bf f}_i+{\bf t}_1)\ \ \textrm{ and }\ \  
{\bf h}_{2i}=y_{i}-s({\bf R}_2\bar{\bf f}_i+{\bf t}_2).$$
We solve ${\bf C}{\bm \alpha}={\bf h}$ in a least squares sense subject to an additional constraint to ensure plausibility of the solution. We follow Brunton \etal \cite{Brunton:14} and use a hyperbox constraint on the shape parameters. This avoids having to choose a regularisation weight but ensures that each parameter lies within $k$ standard deviations of the mean by introducing a linear inequality constraint on the shape parameters (we use $k=3$ in our experiments). Hence, the problem can be solved in closed form as an inequality constrained linear least squares problem.

\subsection{Nonlinear Refinement}

Having alternated pose and shape estimation for a fixed number of iterations, finally we perform nonlinear optimisation of $E_{\textrm{lmk}}$ over ${\bm \alpha}$, ${\bf R}$, ${\bf t}$ and $s$ simultaneously. We represent ${\bf R}$ in axis-angle space to ensure that it remains a valid rotation matrix and we retain the hyperbox constraint on ${\bm \alpha}$. We minimise $E_{\textrm{lmk}}$ using the trust-region-reflective algorithm \cite{Coleman:96} as implemented in the Matlab {\tt lsqnonlin} function.

\section{Fitting with Hard Edge Correspondence}\label{sec:hard}

The method in Section \ref{sec:fitting} enables a 3DMM to be fitted to 2D landmark positions if the correspondence between landmarks and model vertices is known. Edges, for example caused by occluding boundaries, do not have a fixed correspondence to model vertices. Hence, fitting to edges requires shape and pose estimation to happen in conjunction with establishing correspondence between image and model edges. Our proposed approach establishes these correspondences explicitly by finding the closest image edge to each model boundary vertex (subject to additional filtering to remove unreliable matches). Our method comprises the following steps:
\begin{enumerate}
    \item Detect facial landmarks
    \item Initialise shape and pose estimates by fitting to landmarks only
    \item Improve initialisation using iterated closest edge fitting
    \item Nonlinear optimisation of hybrid objective function containing landmark, edge and prior terms
\end{enumerate}
We describe each of these steps in more detail in the rest of this section.

\subsection{Landmarks}\label{sec:landmarksonly}

We use landmarks both for initialisation and as part of our overall objective function as one cue for shape estimation. We apply a facial landmark detector that is suitable for operating on ``in the wild'' images. This provides approximate positions of facial landmarks for which we know the corresponding vertices in the morphable model. We use these landmark positions to make an initial estimate of the pose and shape parameters by running the method in Section \ref{sec:fitting} with only these corresponding landmark positions. Note that any facial landmark detector can be used at this stage. In our experiments, we show results with a recent landmark detection algorithm~\cite{zhu2012face} that achieves state-of-the-art performance and for which code is provided by the authors. In our experimental evaluation, we include the results of fitting to landmarks only.

\subsection{Edge Cost}\label{sec:edgecost}

We assume that a subset of pixels have been labelled as edges and stored as the set ${\cal E}=\{(x,y)|(x,y) \textrm{ is an edge}\}$. In practice, we compute edges by applying the Canny edge detector with a fixed threshold to the input image.

Model contours are computed based on the pose and shape parameters as the occluding boundary of the 3D face. The set of occluding boundary vertices, ${\cal B}({\bm \alpha},{\bf R},{\bf t},s)$, are defined as those lying on a mesh edge whose adjacent faces have a change of visibility. This definition encompasses both outer (silhouette) and inner (self-occluding) contours. Since the viewing direction is aligned with the $z$-axis, this is tested simply by checking if the sign of the $z$-component of the triangle normal changes on either side of the edge. In addition, we check that potential edge vertices are not occluded by another part of the mesh (using z-buffering) and we ignore edges that lie on a mesh boundary since they introduce artificial edges. In this paper, we deal only with occluding contours (both inner and outer). If texture contours were defined on the surface of the morphable model, it would be straightforward to include these in our approach. 

We define the objective function for edge fitting with hard correspondence as the sum of squared distances between each projected occluding boundary vertex and the closest edge pixel:
\begin{align}
&E_{\textrm{edge}}({\bm \alpha},{\bf R},{\bf t},s) = \\
& \frac{1}{|{\cal B}({\bm \alpha},{\bf R},{\bf t},s)|} \sum_{i \in {\cal B}({\bm \alpha},{\bf R},{\bf t},s) } \min_{(x,y) \in {\cal E}} \| [x\ y]^T - \textrm{\bf SOP}\left[{\bf P}_i{\bm \alpha}+\bar{\bf f}_i,{\bf R},{\bf t},s\right] \|^2. \notag
\end{align}
Note that the minimum operator is responsible for computing the hard correspondences. This objective is non-convex since the minimum of a set of convex functions is not convex \cite{grant2006disciplined}. 
Hence, we require a good initialisation to ensure convergence to a minimum close to the global optimum. Fitting to landmarks only does not provide a sufficiently good initialisation. So, in the next subsection we describe a method for obtaining a good initial fit to edges, before incorporating the edge cost into a hybrid objective function in Section \ref{sec:hybrid}.

\subsection{Iterated Closest Edge Fitting}

\begin{figure}[!t]
\centering
\includegraphics[width=4cm, trim=0px 50px 0px 30px, clip=true]{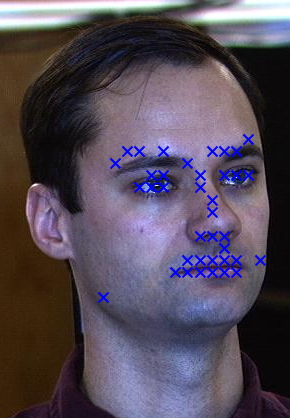}
\includegraphics[width=4cm, trim=0px 50px 0px 30px, clip=true]{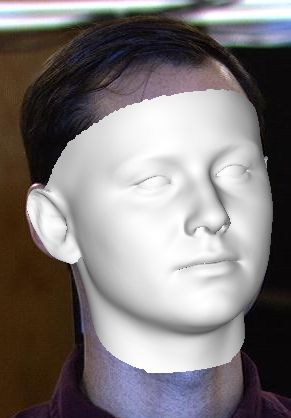}
\includegraphics[width=4cm, trim=0px 161px 0px 97px, clip=true]{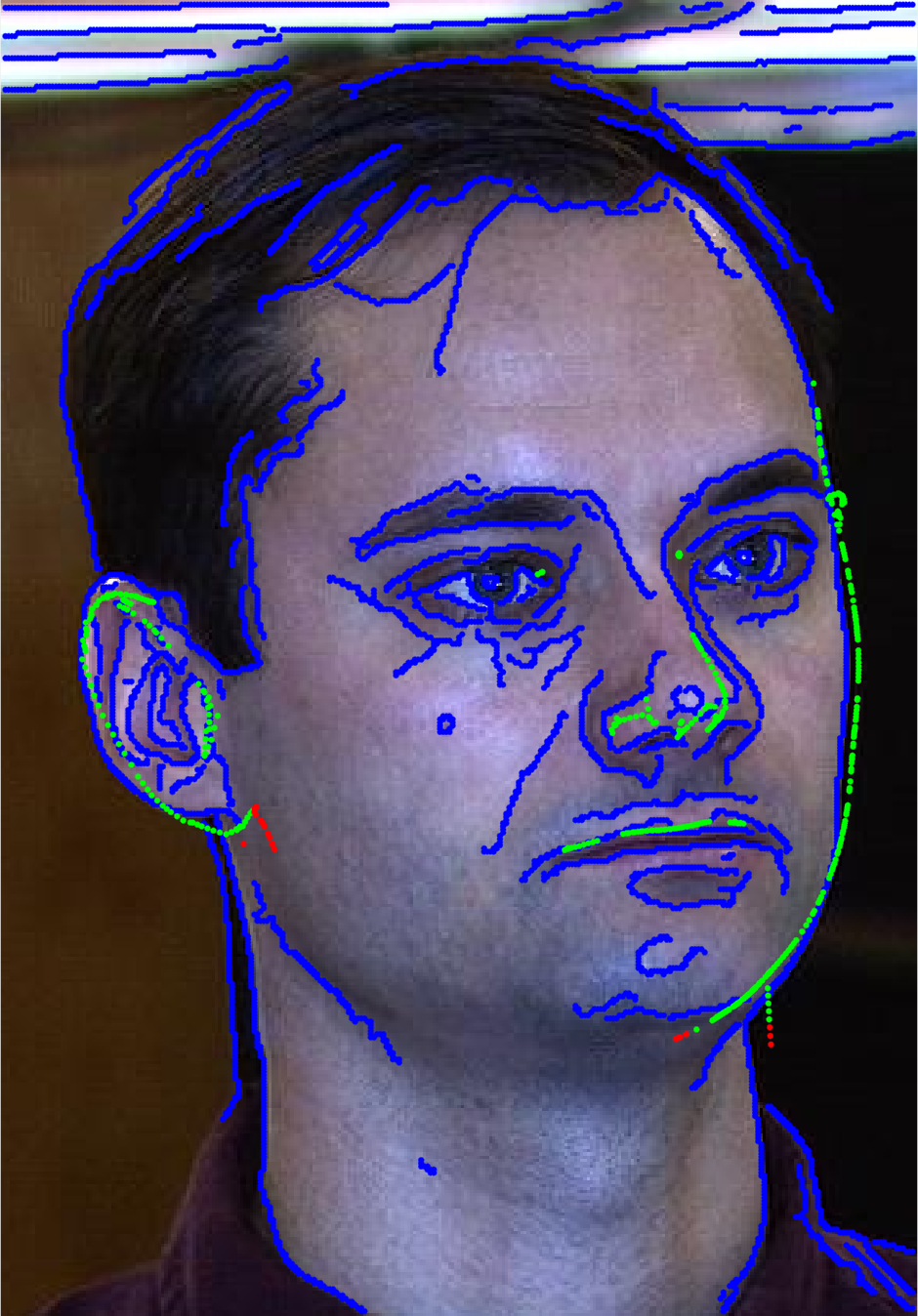}
\caption{Iterated closest edge fitting for initialisation of the edge fitting process. Left: input image with automatically detected landmarks. Middle: overlaid shape obtained by fitting only to landmark. Right: image edges in blue, model boundary vertices with image correspondences in green, unreliable correspondences in red.}
\label{fig:edgefit}
\end{figure}

We propose to refine the landmark-only fit with an initial fit to edges that works in an iterated closest point manner. That is, for each projected model contour vertex, we find the closest image edge pixel and we treat this as a known correspondence. In conjunction with the landmark correspondences, we again run the method in Section \ref{sec:fitting}. This leads to updated pose and shape parameters, and in turn to updated model edges and correspondences. We iterate this process for a fixed number of iterations. We refer to this process as Iterated Closest Edge Fitting (ICEF) and provide an illustration in Figure \ref{fig:edgefit}. On the left we show an input image with the initial landmark detection result. In the middle we show the initial shape and pose obtained by fitting only to landmarks. On the right we show image edge pixels in blue and projected model contours in green (where nearest neighbour edge correspondence is considered reliable) and in red (where correspondence is considered unreliable). The green/blue correspondences are used for the next iteration of fitting.

Finding the image edge pixel closest to a projected contour vertex can be done efficiently by storing the image edge pixels in a $k$d-tree. We filter the resulting correspondences using two commonly used heuristics. First, we remove $5\%$ of the matches for which the distance to the closest image edge pixel is largest. Second, we remove matches for which the image distance divided by $s$ exceeds a threshold (chosen as $10$ in our experiments). The division by scale factor $s$ makes this choice invariant to changes in image resolution.

\subsection{Prior}

Under the assumption that the training data of the 3DMM forms a Gaussian cloud in high dimensional space, then we expect that each of the shape parameters follows a normal distribution with zero mean and variance given by the eigenvalue, $\lambda_i$, associated with the corresponding principal component. We find that including a prior term that captures this assumption significantly improves performance over using the hyperbox constraint alone. The prior penalises deviation from the mean shape as follows:
\begin{equation}
E_{\textrm{prior}}({\bm \alpha}) =\sum_{i=1}^S \left(\frac{\alpha_i}{\sqrt{\lambda_i}}\right)^2.
\end{equation}

\subsection{Nonlinear Refinement}\label{sec:hybrid}

Finally, we perform nonlinear optimisation of a hybrid objective function comprising landmark, edge and prior terms:
\begin{equation}
E({\bm \alpha},{\bf R},{\bf t},s) = w_1E_{\textrm{lmk}}({\bm \alpha},{\bf R},{\bf t},s) + w_2E_{\textrm{edge}}({\bm \alpha},{\bf R},{\bf t},s) + w_3E_{\textrm{prior}}({\bm \alpha}),
\end{equation}
where $w_1$, $w_2$ and $w_3$ weight the contribution of each term to the overall energy. 
The landmark and edge terms are invariant to the number of landmarks and edge vertices which means we do not have to tune the weights for each image (for example, for the results in Table \ref{tab:quangtlandmarks} we use fixed values of: $w_1=0.15$, $w_2=0.45$ and $w_3=0.4$). We retain the hyperbox constraint and so the hybrid objective is a constrained nonlinear least squares problem and we again optimise using the trust-region-reflective algorithm.

For efficiency and to avoid problems of continuity and differentiability of the edge cost function, we follow \cite{Amberg2007} and keep occluding boundary vertices, ${\cal B}$, fixed for a number of iterations of the optimiser. After a number of iterations, we recompute the vertices lying on the occluding boundary and restart the optimiser.

\section{Fitting with Soft Edge Correspondence}\label{sec:soft}

\begin{figure}[!t]
\centering
\includegraphics[width=3cm, trim=0px 50px 0px 30px, clip=true]{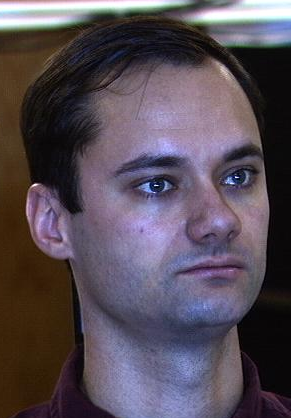}
\includegraphics[width=3cm, trim=0px 50px 0px 30px, clip=true]{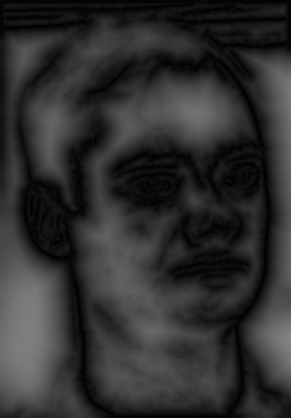}
\caption{Edge cost surface with soft correspondence (right) computed from input image (left)}
\label{fig:DT}
\end{figure}

We compare our approach with a method based on optimising an edge cost function, in the same spirit as previous work \cite{Amberg2007,keller20073d,Romdhani:05}.
We follow the same approach as Amberg~\etal \cite{Amberg2007} to compute the edge cost function, however we further improve robustness by also integrating over scale. 
For our edge detector, we use gradient magnitude thresholding with non-maxima suppression.
Given a set of edge detector sensitivity thresholds $\cal{T}$ and scales $\cal{S}$, we compute $n=|\cal{T}\times\cal{S}|$ edge images, $E^1$, $\dots$, $E^n$, using each pair of image scale and threshold values. We compute the Euclidean distance transform, $D^1$, $\dots$, $D^n$, for each edge image (i.e. the value of each pixel in $D^i$ is the distance to the closest edge pixel in $E^i$). Finally, we compute the edge cost surface as:
\begin{equation}
 S(x,y) = \frac{1}{n}\sum_{i=1}^n \frac{D^i(x,y)}{D^i(x,y)+\kappa}.
\end{equation}
The parameter $\kappa$ determines the influence range of an edge in an adaptive manner. Amberg~\etal \cite{Amberg2007} suggest a value for $\kappa$ of 1/20th the expected size of the head in pixels. We compute this parameter automatically from the scale $s$. An example of an edge cost surface is shown in Figure \ref{fig:DT}. To evaluate the edge cost, we compute model contour vertices as in Section \ref{sec:edgecost}, project them into the image and interpolate the edge cost function using bilinear interpolation:
\begin{equation}
 E_{\textrm{softedge}}({\bm \alpha},{\bf R},{\bf t},s) = \frac{1}{|{\cal B}({\bm \alpha},{\bf R},{\bf t},s)|} \sum_{i \in {\cal B}({\bm \alpha},{\bf R},{\bf t},s) } S(\textrm{\bf SOP}\left[{\bf P}_i{\bm \alpha}+\bar{\bf f}_i,{\bf R},{\bf t},s\right]).
\end{equation}

As with the hard edge cost, we found that the best performance was achieved by also including the landmark and prior terms in a hybrid objective function. Hence, we minimise:
\begin{equation}
E({\bm \alpha},{\bf R},{\bf t},s) = w_1E_{\textrm{lmk}}({\bm \alpha},{\bf R},{\bf t},s) + w_2E_{\textrm{softedge}}({\bm \alpha},{\bf R},{\bf t},s) + w_3E_{\textrm{prior}}({\bm \alpha}).
\end{equation}
We again initialise by fitting to landmarks only using the method in Section \ref{sec:landmarksonly}, retain the hyperbox constraint and optimise using the trust-region-reflective algorithm. We use the same weights as for the hard correspondence method in our experiments.

\begin{figure}[!t]
\centering
\includegraphics[height=1.5cm, trim=181px 186px 93px 208px, clip=true]{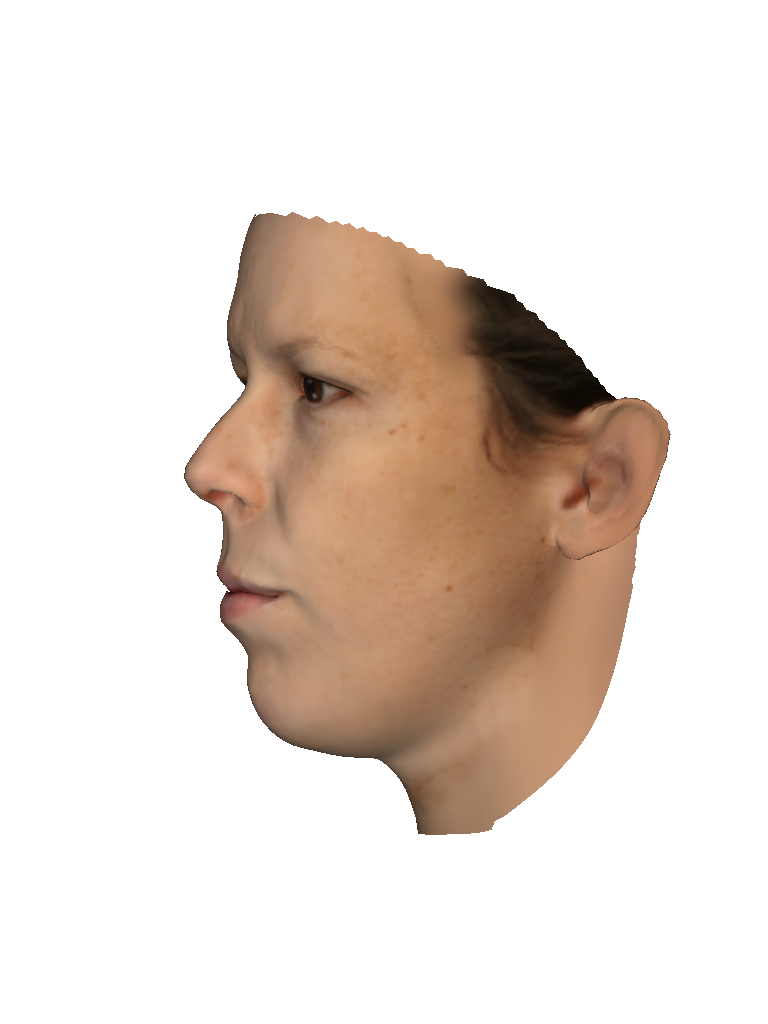}
\includegraphics[height=1.5cm, trim=210px 186px 50px 208px,clip=true]{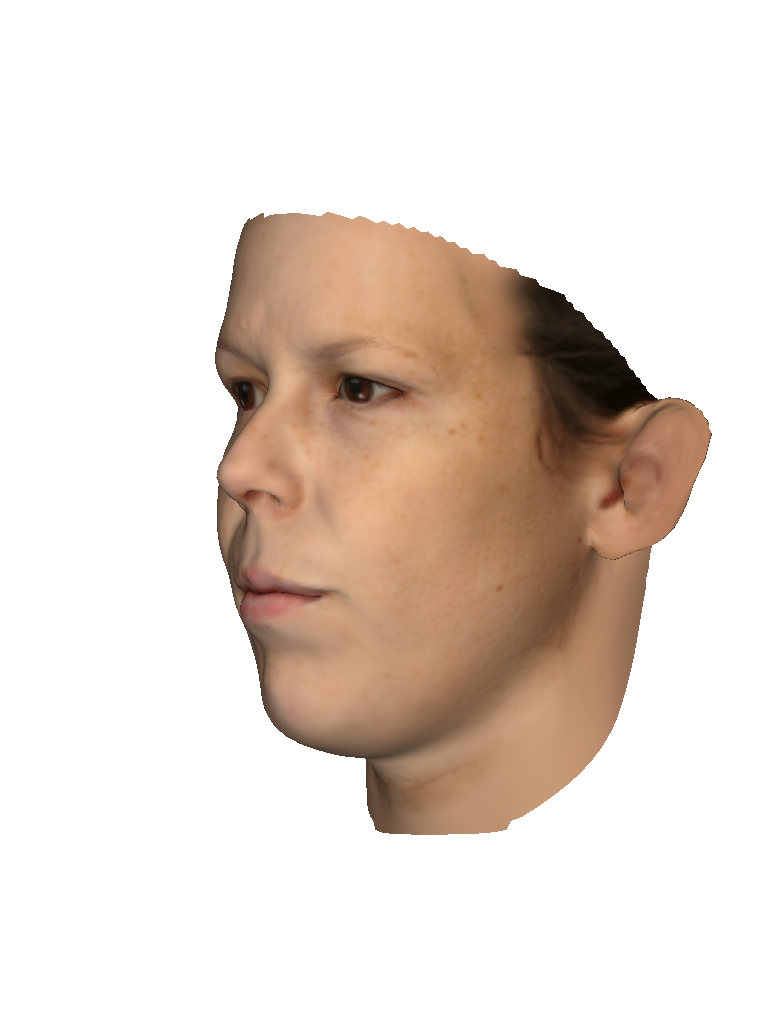}
\includegraphics[height=1.5cm, trim=201px 186px 50px 208px,clip=true]{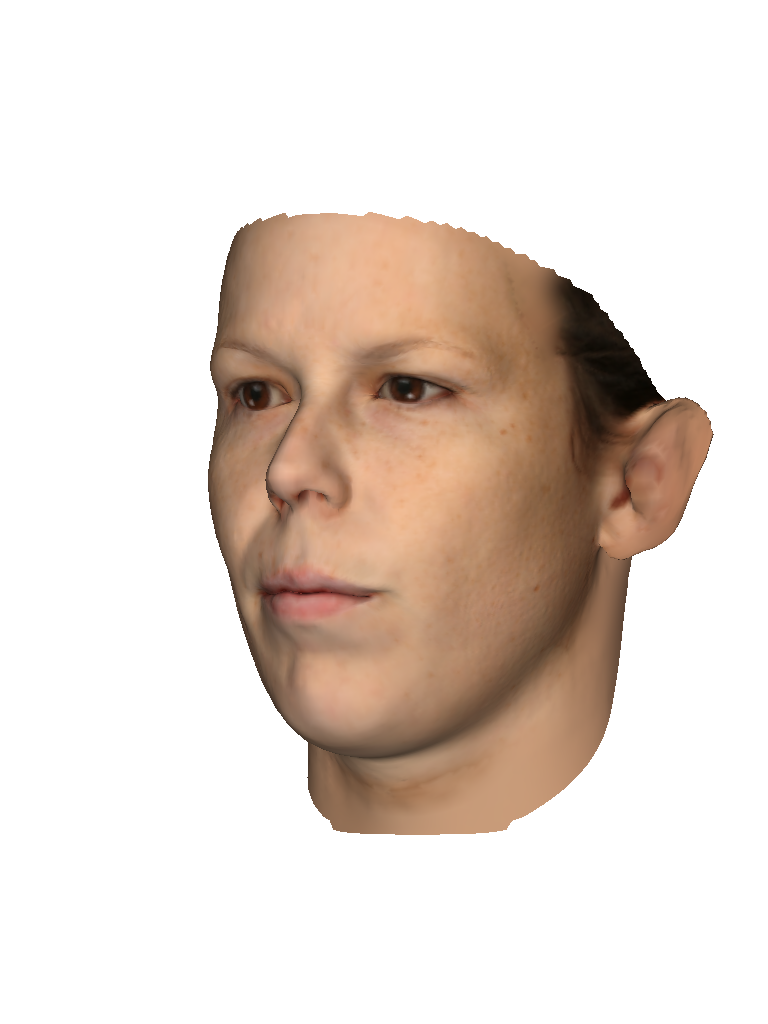}
\includegraphics[height=1.5cm, trim=178px 186px 74px 208px,clip=true]{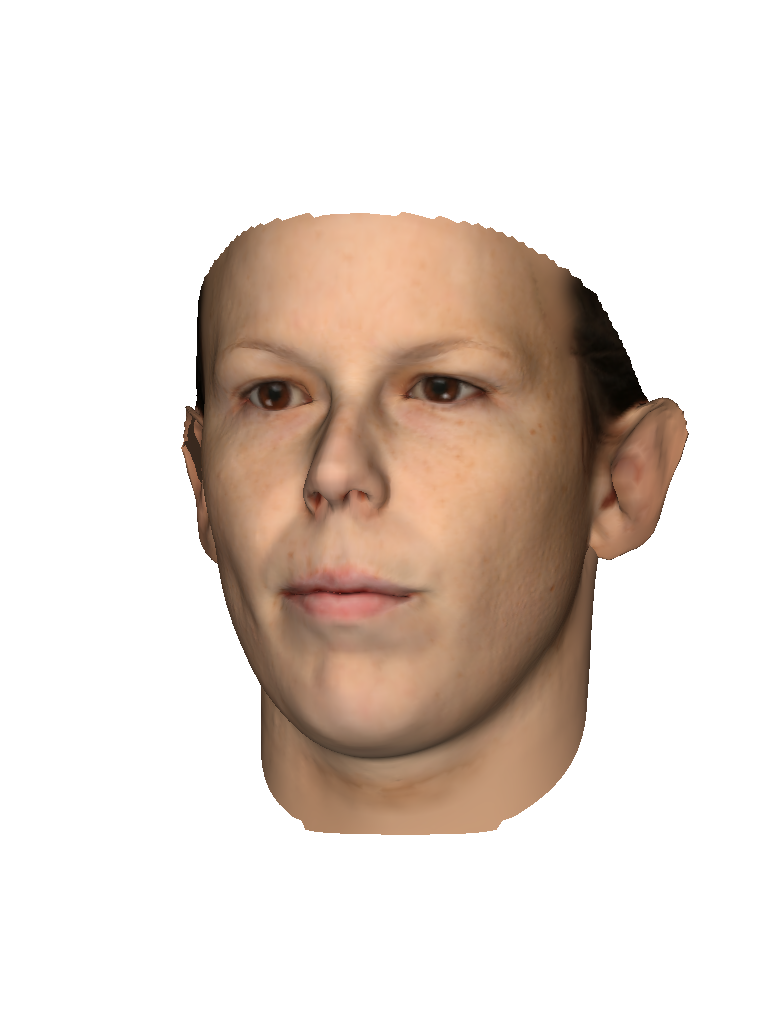}
\includegraphics[height=1.5cm, trim=117px 186px 122px 208px,clip=true]{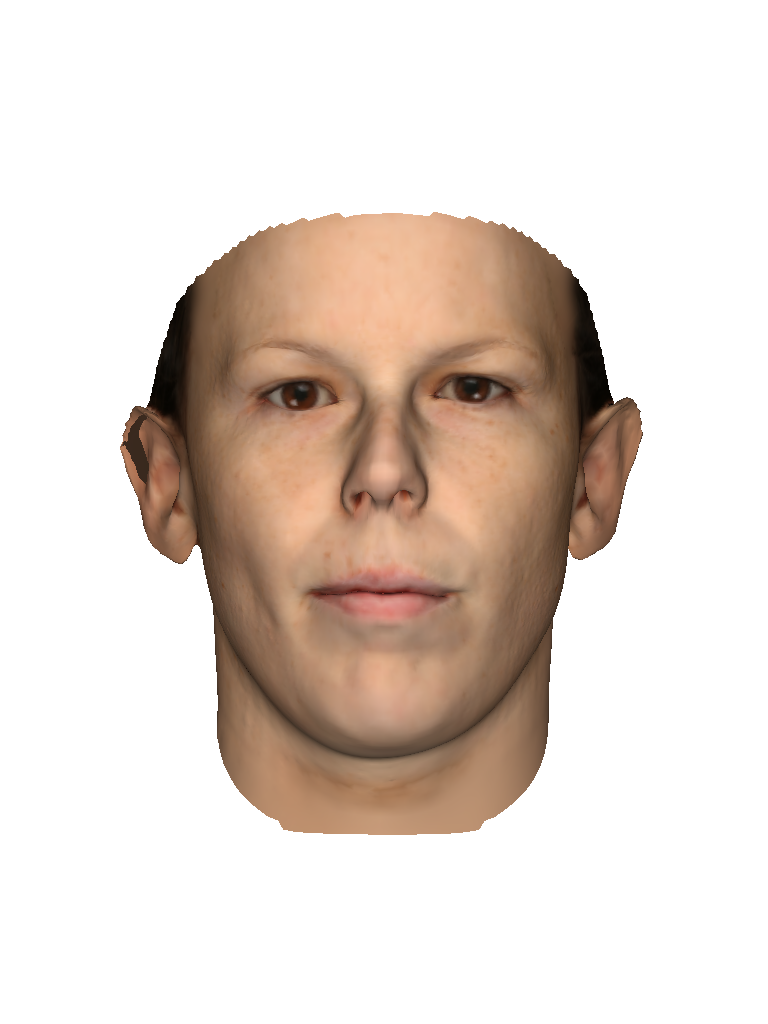}
\includegraphics[height=1.5cm, trim=76px 186px 182px 208px,clip=true]{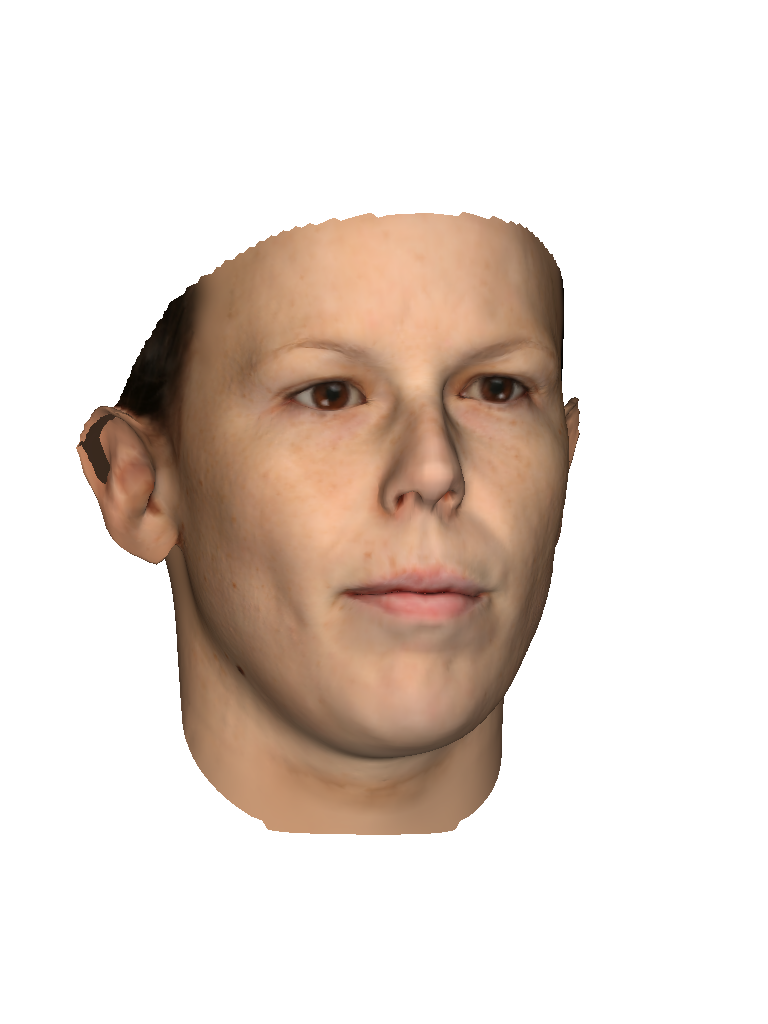}
\includegraphics[height=1.5cm, trim=52px 186px 200px 208px,clip=true]{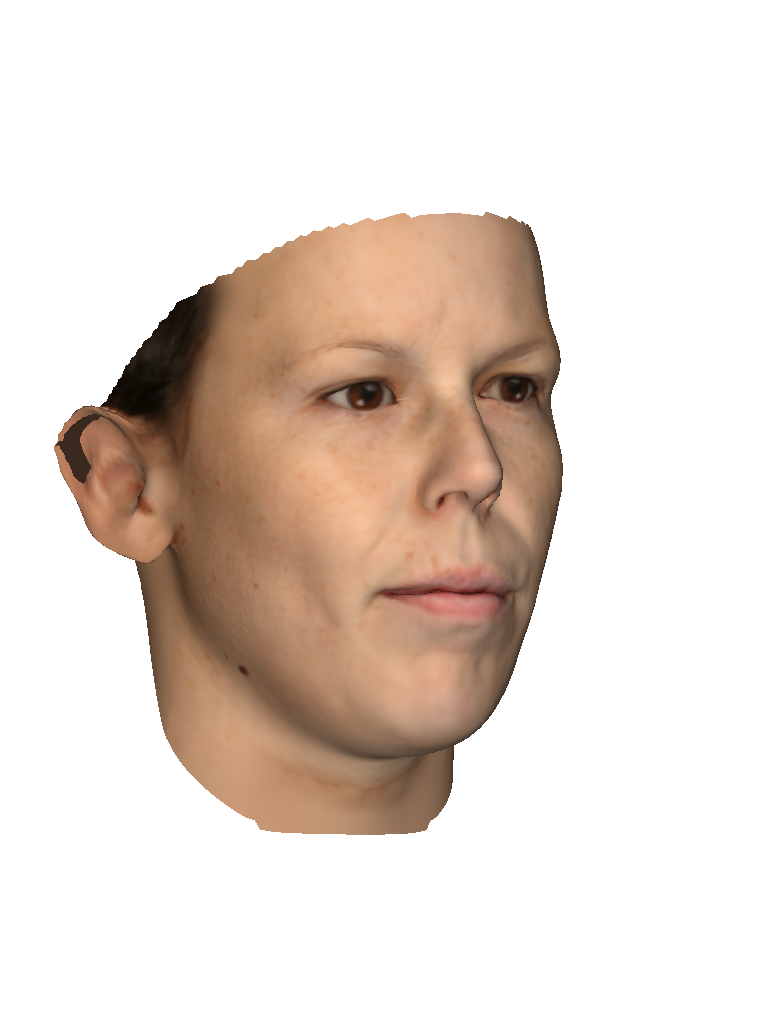}
\includegraphics[height=1.5cm, trim=55px 186px 210px 208px,clip=true]{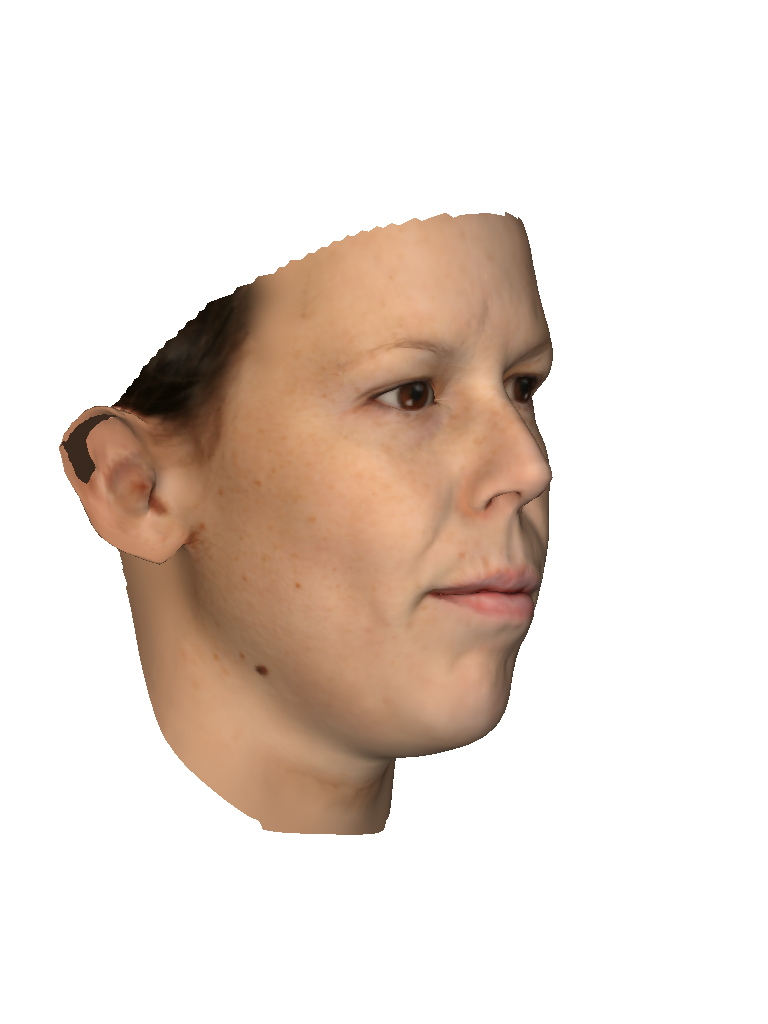}
\includegraphics[height=1.5cm, trim=98px 186px 182px 208px,clip=true]{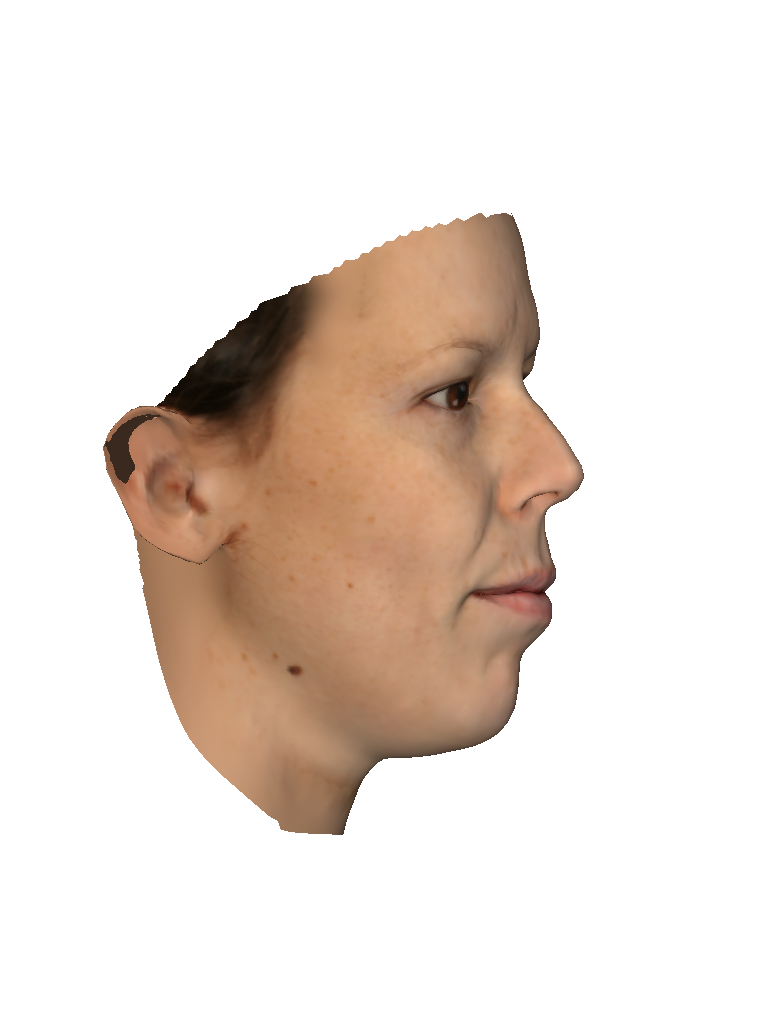}
\caption{Synthetic input images for one subject}
\label{fig:input}
\end{figure}

\begin{table}[!t]
\centering
\begin{tabular}{|l||c|c|c|c|c|c|c|c|c|c|}\hline

 & \multicolumn{9}{c|}{{\bf Rotation angle}} &  \\ \cline{2-10}

{\bf Method} & $-70^{\circ}$ & $-50^{\circ}$ & $-30^{\circ}$ & $-15^{\circ}$ & $0^{\circ}$ & $15^{\circ}$ & $30^{\circ}$ & $50^{\circ}$ & $70^{\circ}$ & {\bf Mean} \\
\hline
\hline
Average face&3.35&3.35&3.35&3.35&3.35&3.35&3.35&3.35&3.35&3.35\\ \hline
Proposed (landmarks only)&2.67&2.60&2.58&2.64&2.56&2.49&2.50&2.54&2.63&2.58\\ \hline
Aldrian and Smith \cite{aldrian2013inverse}&2.64&2.60&2.55&2.54&{\bf 2.49}&2.42&2.43&2.44&2.54&2.52\\ \hline
Romdhani \etal \cite{Romdhani:05} (soft)&2.65&2.59&2.58&2.61&2.59&2.50&2.50&2.46&2.51&2.55\\ \hline
Proposed (ICEF)&2.38&2.40&2.51&{\bf 2.38}&2.52&2.45&2.43&2.38&2.3&2.42\\ \hline
Proposed (hard)&{\bf 2.35}&{\bf 2.26}&{\bf 2.38}&2.40&2.51&{\bf 2.39}&{\bf 2.40}&{\bf 2.20}&{\bf 2.26}&{\bf 2.35}\\ \hline
\end{tabular}
\caption{Mean Euclidean vertex distance (mm) with ground truth landmarks}
\label{tab:quangtlandmark}
\end{table}

\section{Experimental Results}\label{sec:exp}

We present two sets of experimental results. First, we use synthetic images with known ground truth 3D shape in order to quantitatively evaluate our method and provide comparison to previous work. Second, we use real images to provide qualitative evidence of the performance of our method in uncontrolled conditions.
For the 3DMM in both sets of experiments we use the Basel Face Model \cite{Paysan:09}. 

\subsection{Quantitative Evaluation}

\begin{table}[!t]
\centering
\begin{tabular}{|l||c|c|c|c|c|c|}\hline
 & \multicolumn{6}{c|}{{\bf Landmark noise std. dev.}} \\ \cline{2-7}
{\bf Method} & $\sigma=0$ & $\sigma=1$ & $\sigma=2$ & $\sigma=3$ & $\sigma=4$ & $\sigma=5$ \\
\hline
\hline
Proposed (landmarks only)&2.58&2.60&2.61&2.68&2.76&2.85\\ \hline
Aldrian and Smith \cite{aldrian2013inverse}&2.52&2.53&2.55&2.62&2.65&2.73\\ \hline
Romdhani \etal \cite{Romdhani:05} (soft)&2.55&2.57&2.57&2.62&2.70&2.76\\ \hline
Proposed (ICEF)&2.42&2.43&2.43&2.50&2.57&2.60\\ \hline
Proposed (hard)&{\bf 2.35}&{\bf 2.36}&{\bf 2.35}&{\bf 2.39}&{\bf 2.47}&{\bf 2.50}\\ \hline
\end{tabular}
\caption{Mean Euclidean vertex distance (mm) with noisy landmarks}
\label{tab:quannoisylandmark}
\end{table}

\begin{table}[!t]
\centering
\begin{tabular}{|l|c|c|c|c|c|c|c|c|c|c|}\hline

 & \multicolumn{9}{c|}{{\bf Rotation angle}} &  \\ \cline{2-10}

{\bf Method} & $-70^{\circ}$ & $-50^{\circ}$ & $-30^{\circ}$ & $-15^{\circ}$ & $0^{\circ}$ & $15^{\circ}$ & $30^{\circ}$ & $50^{\circ}$ & $70^{\circ}$ & {\bf Mean} \\
\hline
\hline
Proposed (landmarks only)&6.79&6.84&5.19&5.74&5.68&6.34&6.48&7.04&7.74&6.43\\ \hline
Zhu \etal \cite{Zhu:15}&N/A&N/A&4.63&5.09&4.19&5.22&4.92&N/A&N/A&N/A\\ \hline
Romdhani \etal \cite{Romdhani:05} (soft)&4.46&3.42&3.66&3.78&3.77&3.57&4.31&4.19&4.73&3.99\\ \hline
Proposed (ICEF)&3.70&3.32&3.26&3.23&3.37&3.50&3.43&4.07&3.52&3.49\\ \hline
Proposed (hard)&{\bf3.43}&{\bf 3.20}&{\bf 3.19}&{\bf 3.09}&{\bf 3.30}&{\bf 3.36}&{\bf 3.36}&{\bf 3.84}&{\bf 3.41}&{\bf 3.35}\\ \hline
\end{tabular}
\caption{Mean Euclidean vertex distance (mm) with automatically detected landmarks}
\label{tab:quanlandmarkdetector}
\end{table}

We begin with a quantitative comparative evaluation on synthetic data. We use the 10 out-of-sample faces supplied with the Basel Face Model and render orthographic images of each face in 9 poses (rotations of $0^{\circ}$, $\pm 15^{\circ}$, $\pm 30^{\circ}$, $\pm 50^{\circ}$ and $\pm 70^{\circ}$ about the vertical axis). We show sample input images for one subject in Figure \ref{fig:input}. In all experiments, we report the mean Euclidean distance between ground truth and estimated face surface in mm after Procrustes alignment.

In the first experiment, we use ground truth landmarks. Specifically, we use the 70 Farkas landmarks, project the visible subset to the image (yielding between 37 and 65 landmarks per image) and round to the nearest pixel. In Table \ref{tab:quangtlandmark} we show results averaged over pose angle and over the whole dataset. As a baseline, we show the error if we simply use the average face shape. We then show the result of fitting only to landmarks, i.e. the method in Section \ref{sec:fitting}. We include two comparison methods. The approach of Aldrian and Smith \cite{aldrian2013inverse} uses only landmarks but with an affine camera model and a learnt model of landmark variance. The soft edge correspondence method of Romdhani \etal \cite{Romdhani:05} is described in Section \ref{sec:soft}. The final two rows show two variants of our proposed methods: the fast Iterated Closest Edge Fitting version and the full version with nonlinear optimisation of the hard correspondence cost. Average performance over the whole dataset is best for our method and, in general, using edges over landmarks only and applying nonlinear optimisation improves performance. The performance improvement of our methods over landmark-only methods improves with pose angle. This suggest that edge information becomes more salient for non-frontal poses.

The second experiment is identical to the first except that we add Gaussian noise of varying standard deviation to the ground truth landmark positions. In Table \ref{tab:quannoisylandmark} we show results averaged over all poses and subjects.

In the final experiment we use landmarks that are automatically detected using the method of Zhu and Ramanan \cite{zhu2012face}. This enables us to include comparison with the recent fitting algorithm of Zhu \etal \cite{Zhu:15}. We use the author's own implementation which only works with a fixed set of 68 landmarks. This means that the method cannot be applied to the more extreme pose angles where fewer landmarks are detected. In this more challenging scenario, our method again gives the best overall performance and is superior for all pose angles.

\newcommand{\frontimsize}{3.2cm}
\setlength{\tabcolsep}{1pt}
\renewcommand{\arraystretch}{0.3}
\begin{figure}[!t]
\centering
\begin{tabular}{cccc}
\includegraphics[height=\frontimsize, trim=133px 18px 257px 160px, clip=true]{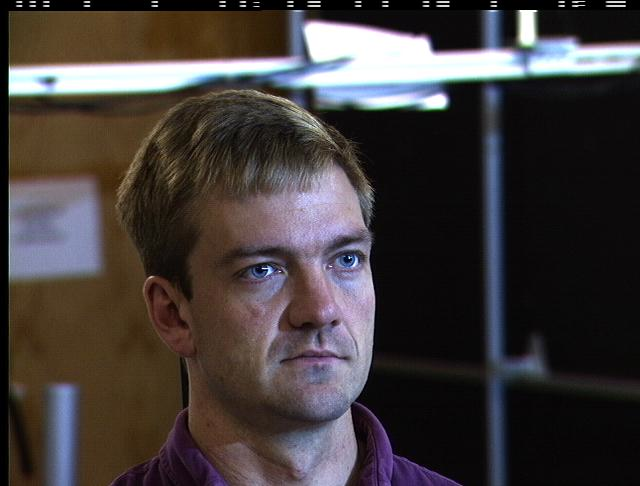}&
\includegraphics[height=\frontimsize, trim=150px 10px 238px 160px, clip=true]{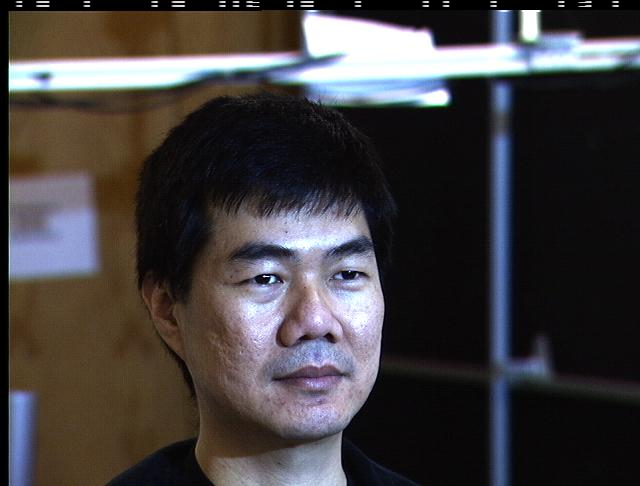}&
\includegraphics[height=\frontimsize, trim=140px 70px 221px 100px, clip=true]{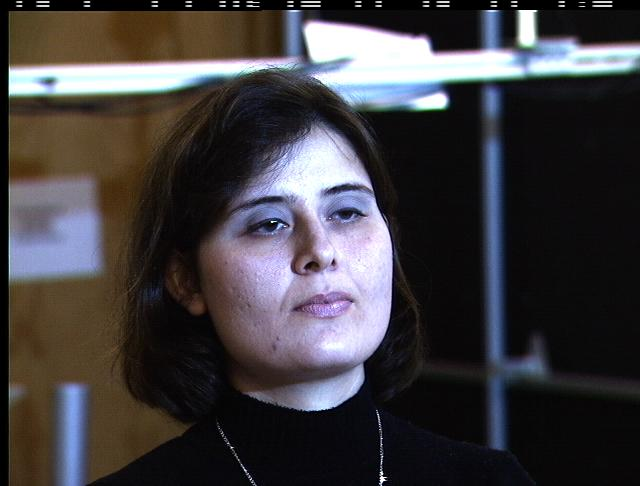}&
\includegraphics[height=\frontimsize, trim=195px 33px 176px 164px, clip=true]{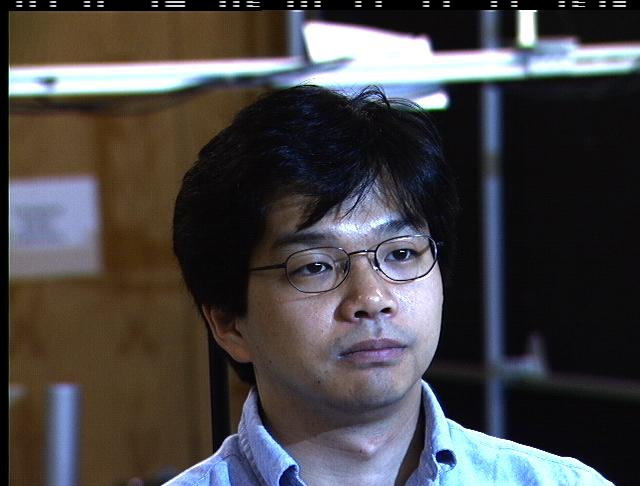}\\
\includegraphics[height=\frontimsize, trim=55px 165px 205px 180px, clip=true]{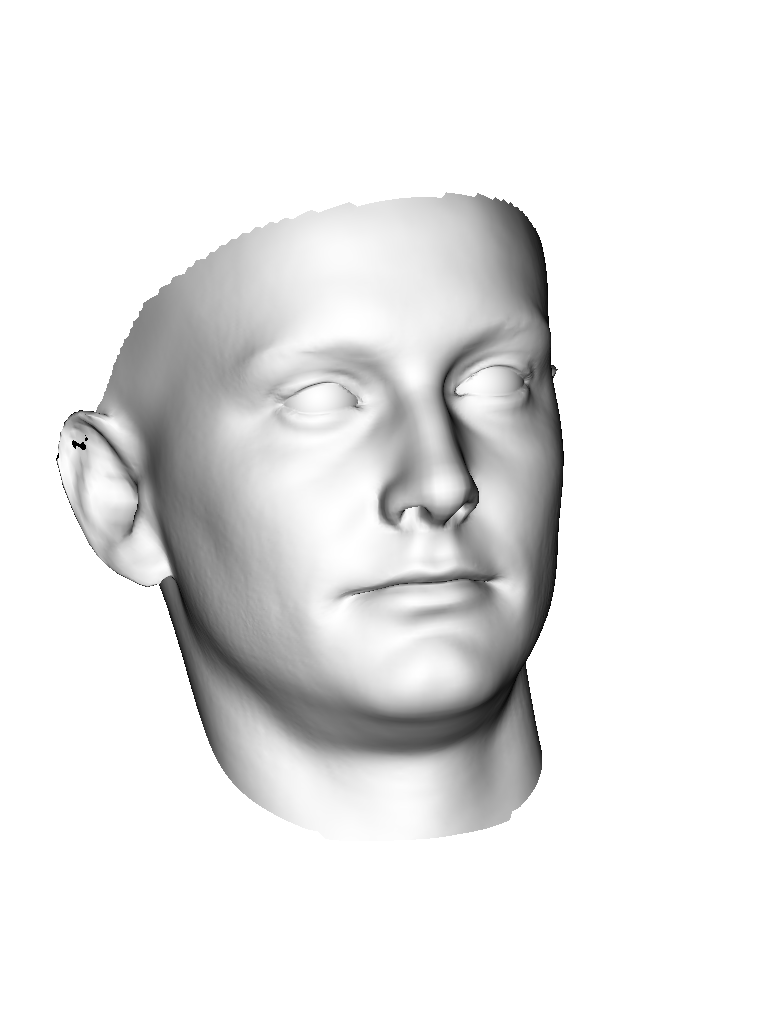}&
\includegraphics[height=\frontimsize, trim=64px 200px 140px 171px, clip=true]{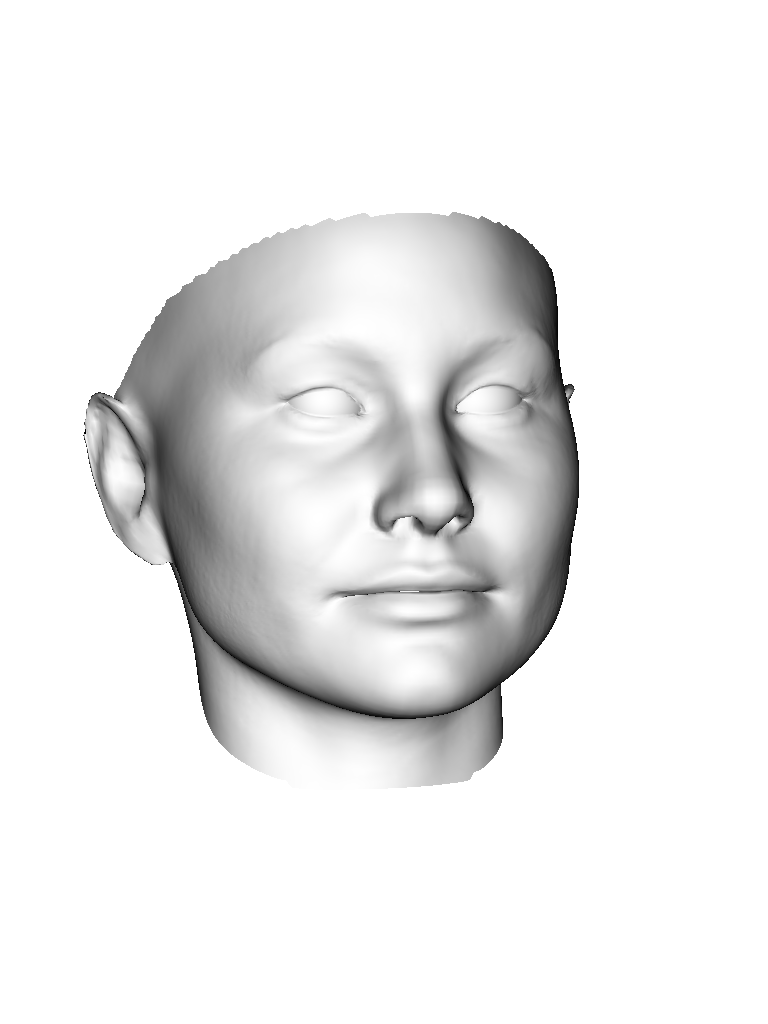}&
\includegraphics[height=\frontimsize, trim=80px 200px 180px 191px, clip=true]{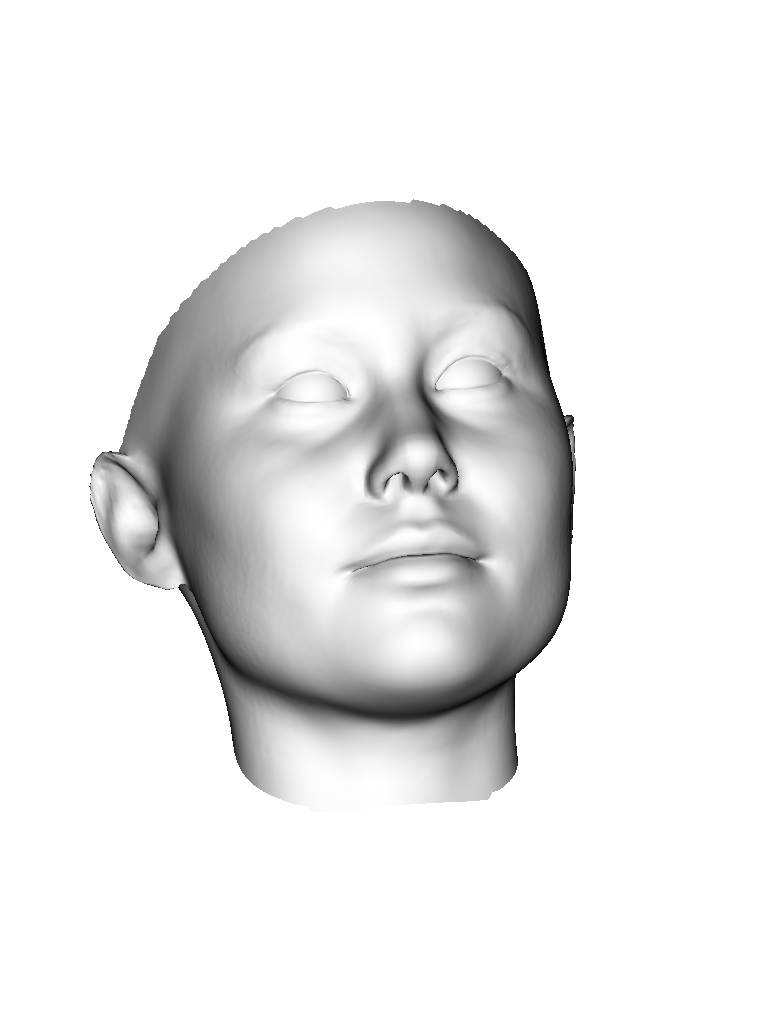}&
\includegraphics[height=\frontimsize, trim=30px 200px 135px 207px, clip=true]{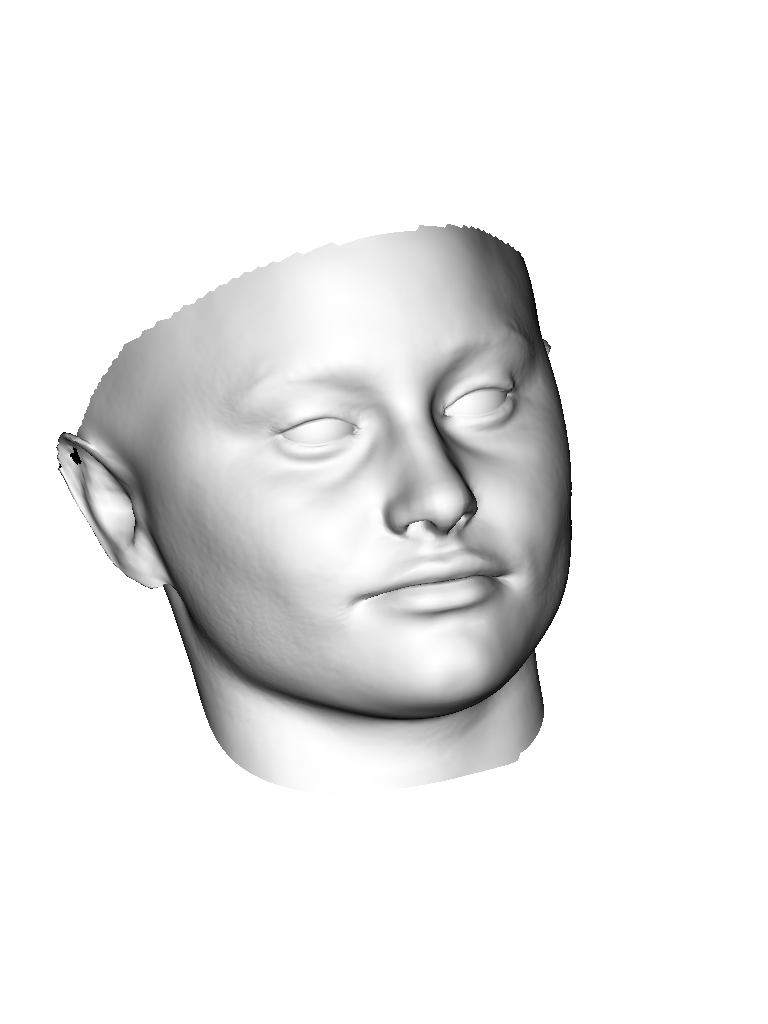}\\
\includegraphics[height=\frontimsize, trim=170px 18px 220px 160px, clip=true]{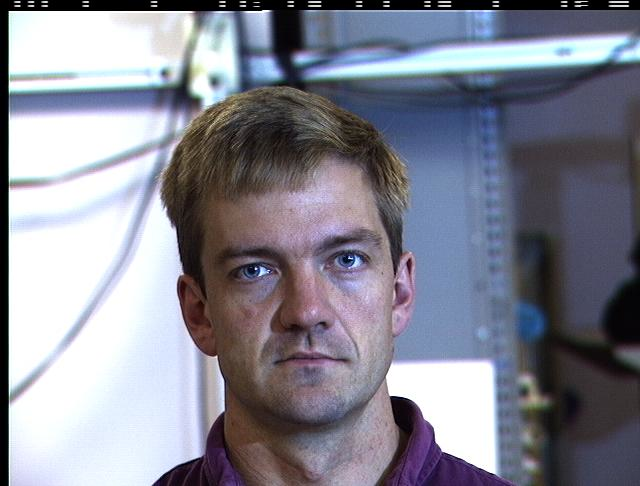}&
\includegraphics[height=\frontimsize, trim=197px 64px 225px 158px, clip=true]{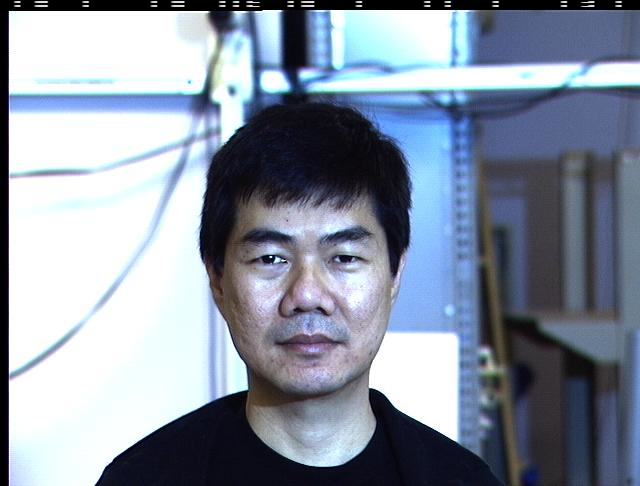}&
\includegraphics[height=\frontimsize, trim=175px 90px 210px 90px, clip=true]{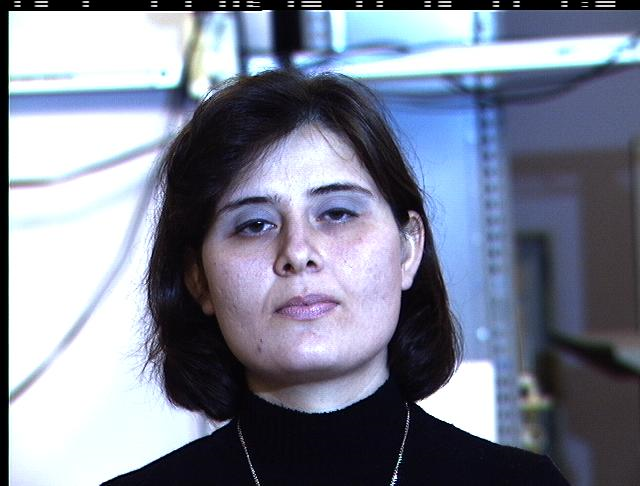}&
\includegraphics[height=\frontimsize, trim=205px 33px 166px 164px, clip=true]{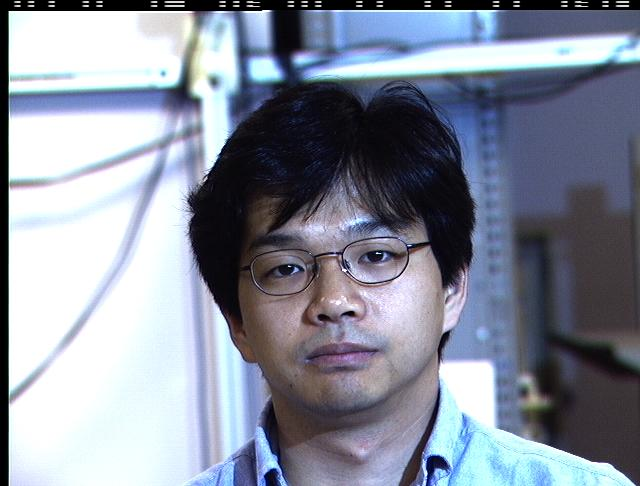}\\
\includegraphics[height=\frontimsize, trim=117px 169px 178px 186px, clip=true]{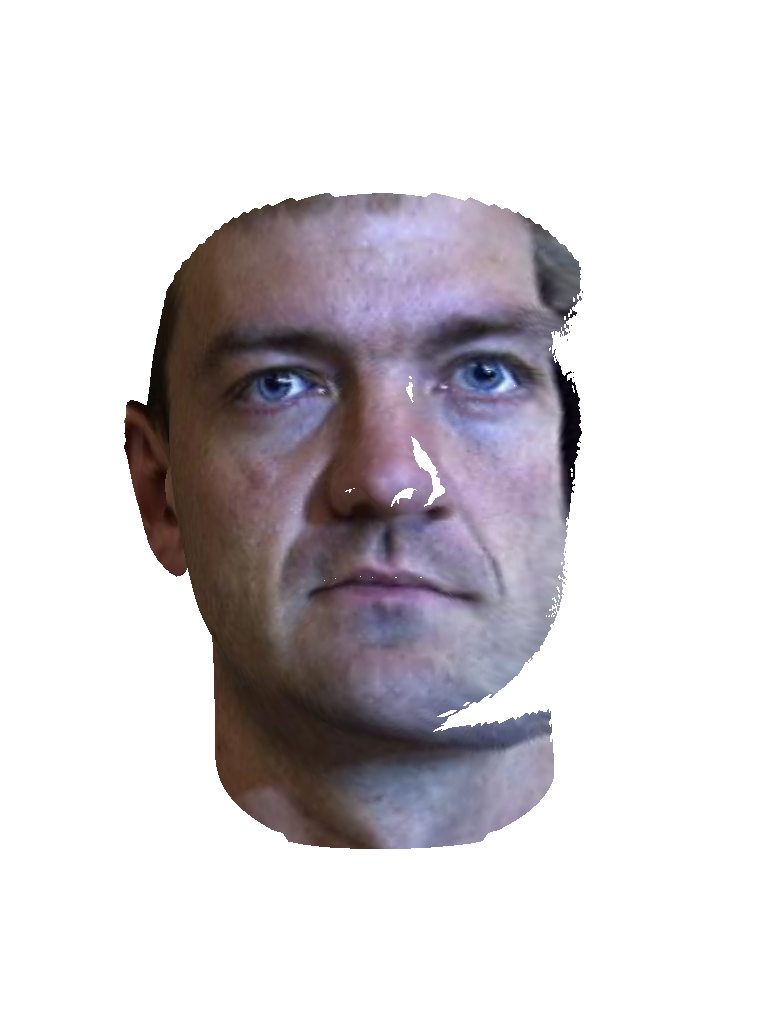}&
\includegraphics[height=\frontimsize, trim=84px 161px 140px 171px, clip=true]{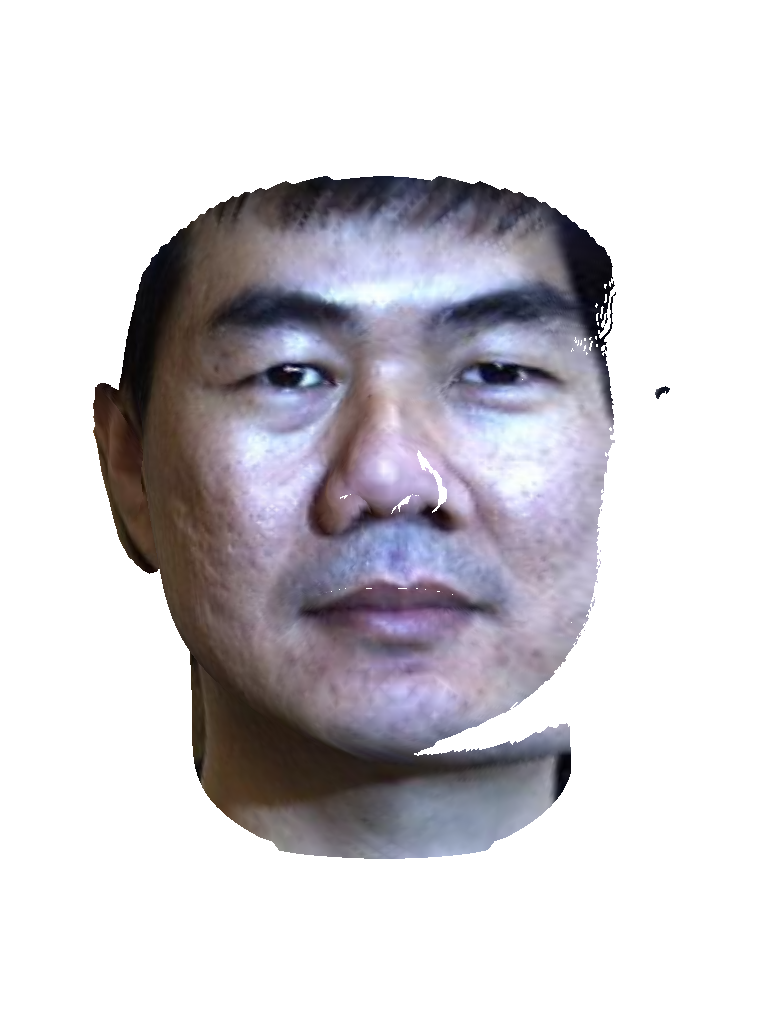}&
\includegraphics[height=\frontimsize, trim=125px 220px 180px 210px, clip=true]{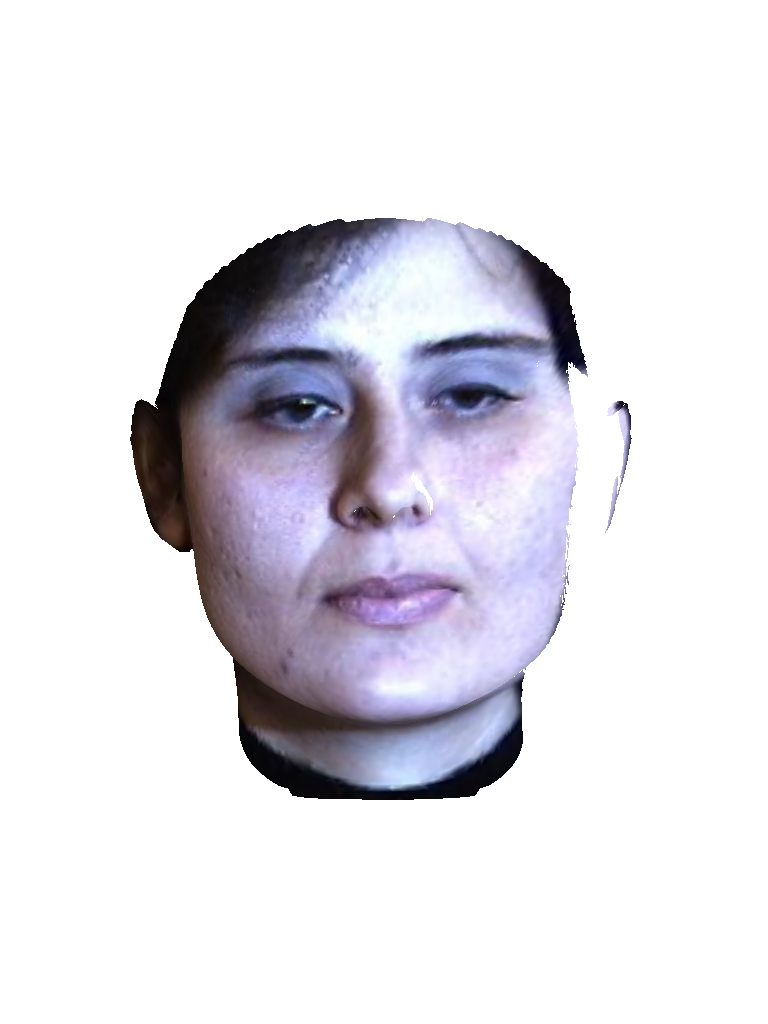}&
\includegraphics[height=\frontimsize, trim=117px 239px 178px 237px, clip=true]{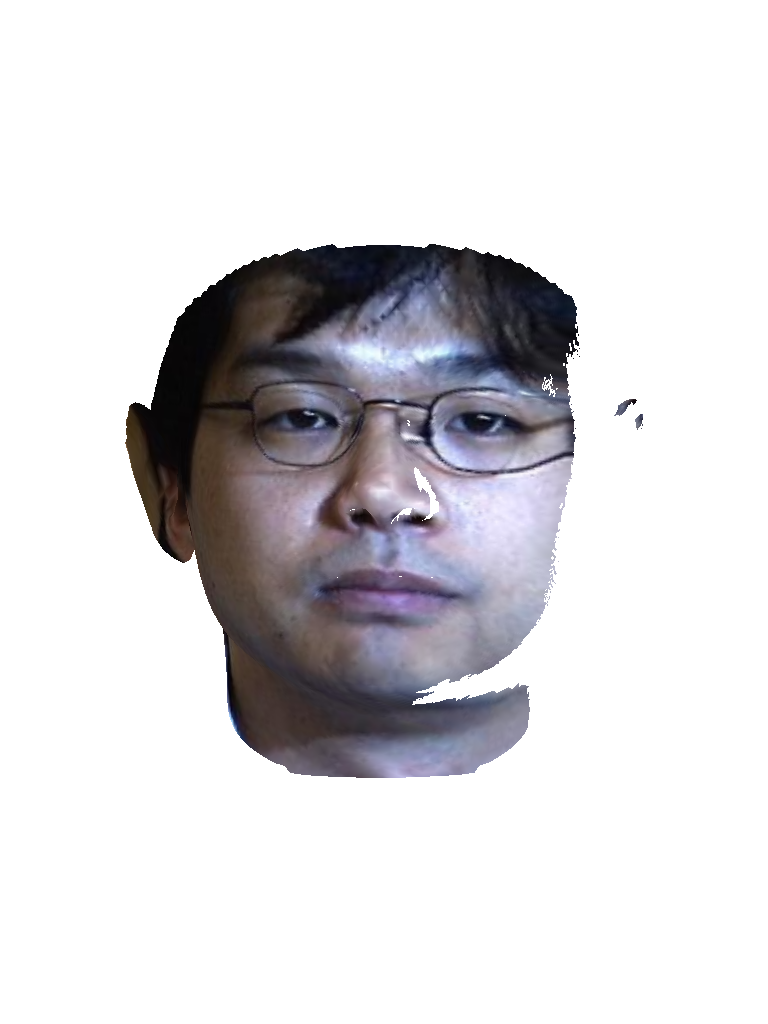}
\end{tabular}
\caption{Qualitative frontalisation results}
\label{fig:CMU}
\end{figure}

\subsection{Qualitative Evaluation}

\begin{figure}[!t]
\centering
\begin{tabular}{cccccc}
\includegraphics[height=2.4cm, trim=60px 46px 57px 30px, clip=true]{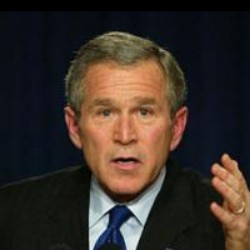}&
\includegraphics[height=2.4cm, trim=140px 265px 125px 135px, clip=true]{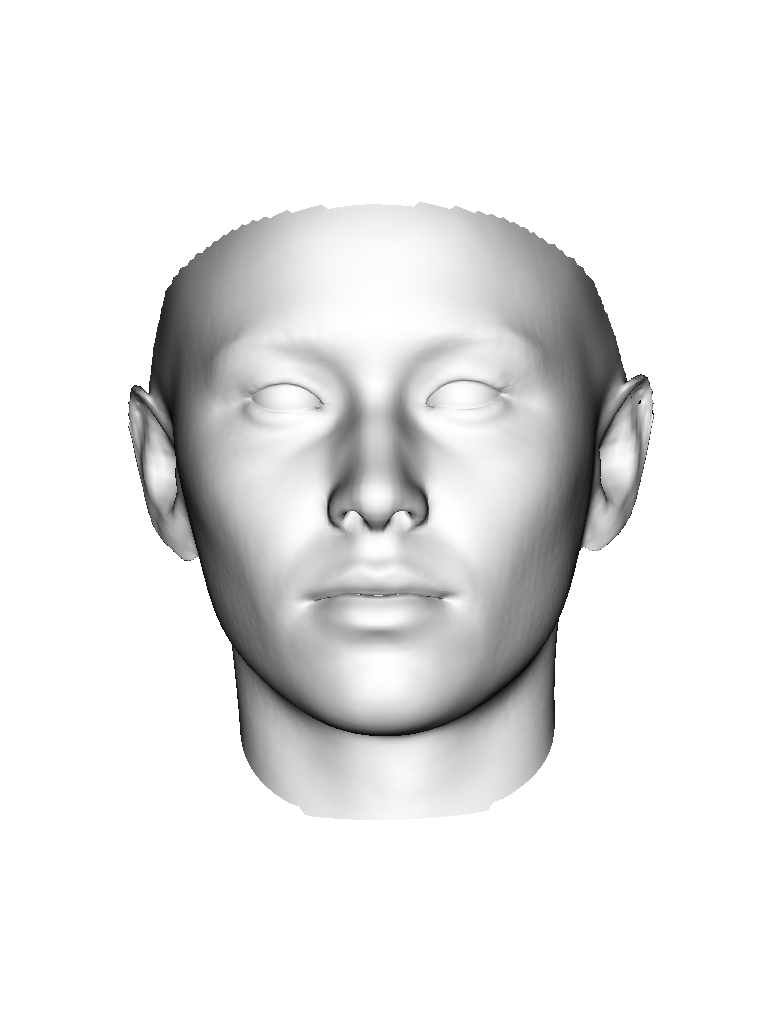}&
\includegraphics[height=2.4cm, trim=194px 193px 58px 194px, clip=true]{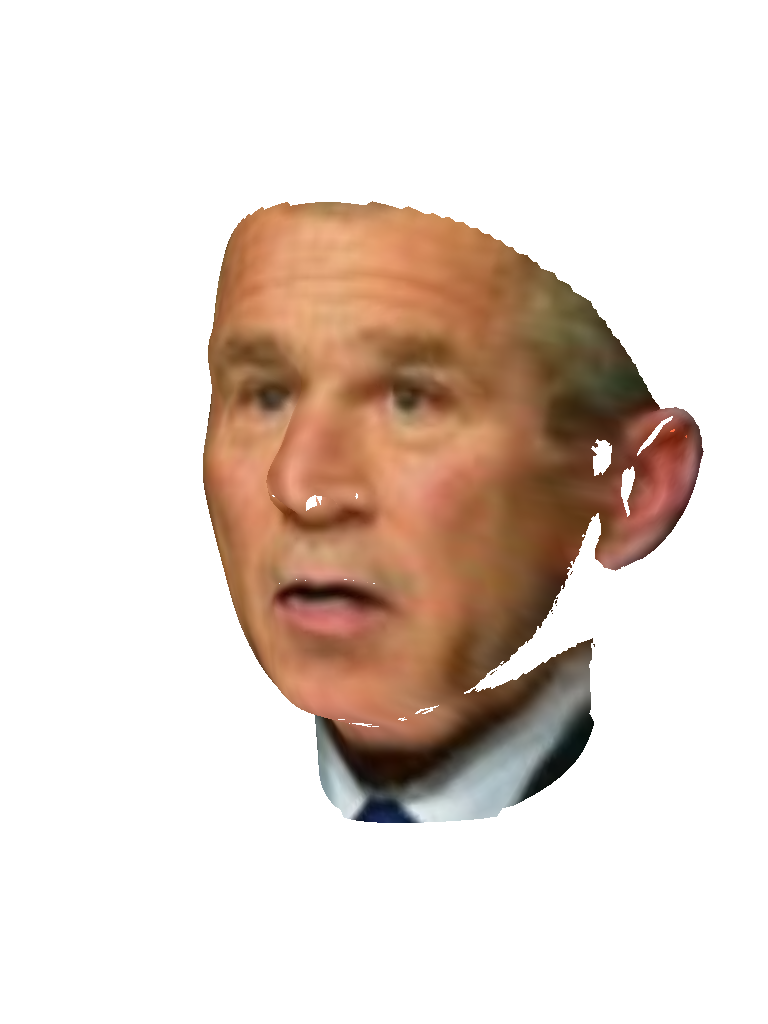}&
\includegraphics[height=2.4cm, trim=171px 193px 75px 194px, clip=true]{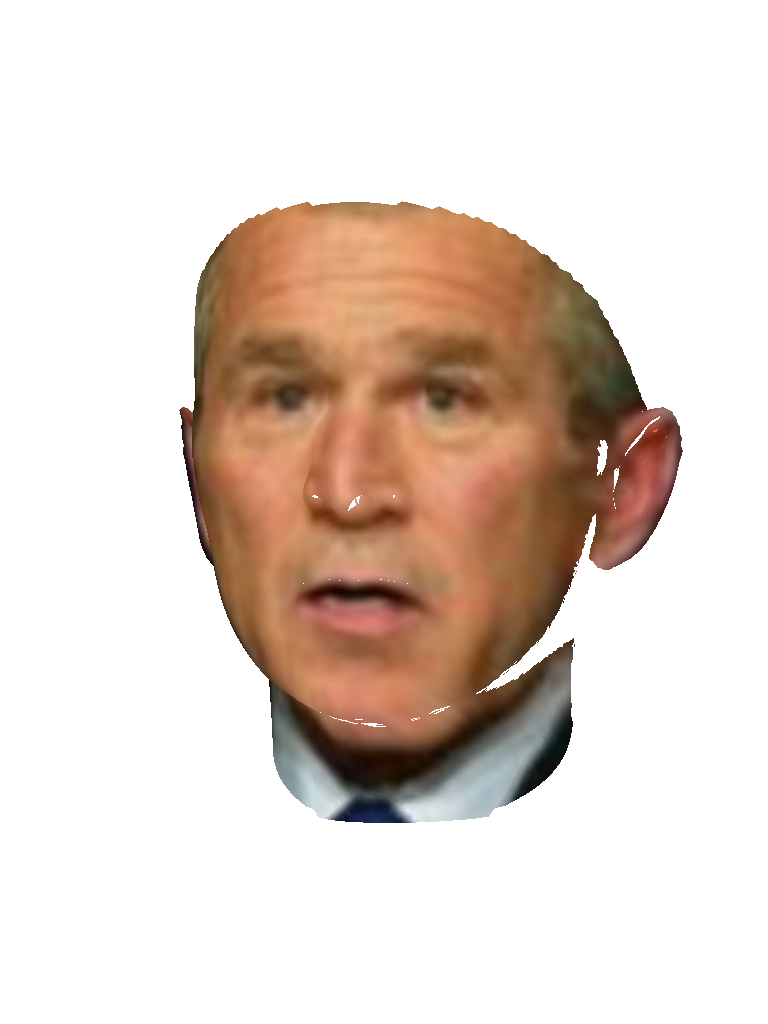}&
\includegraphics[height=2.4cm, trim=81px 193px 172px 194px, clip=true]{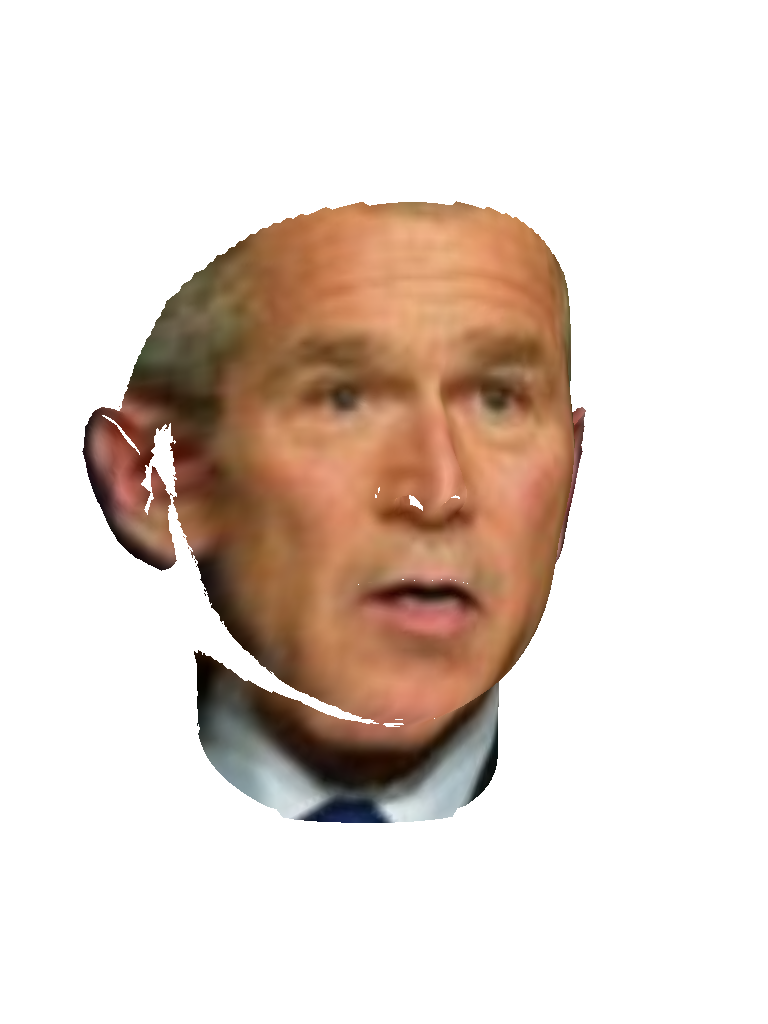}&
\includegraphics[height=2.4cm, trim=56px 193px 194px 194px, clip=true]{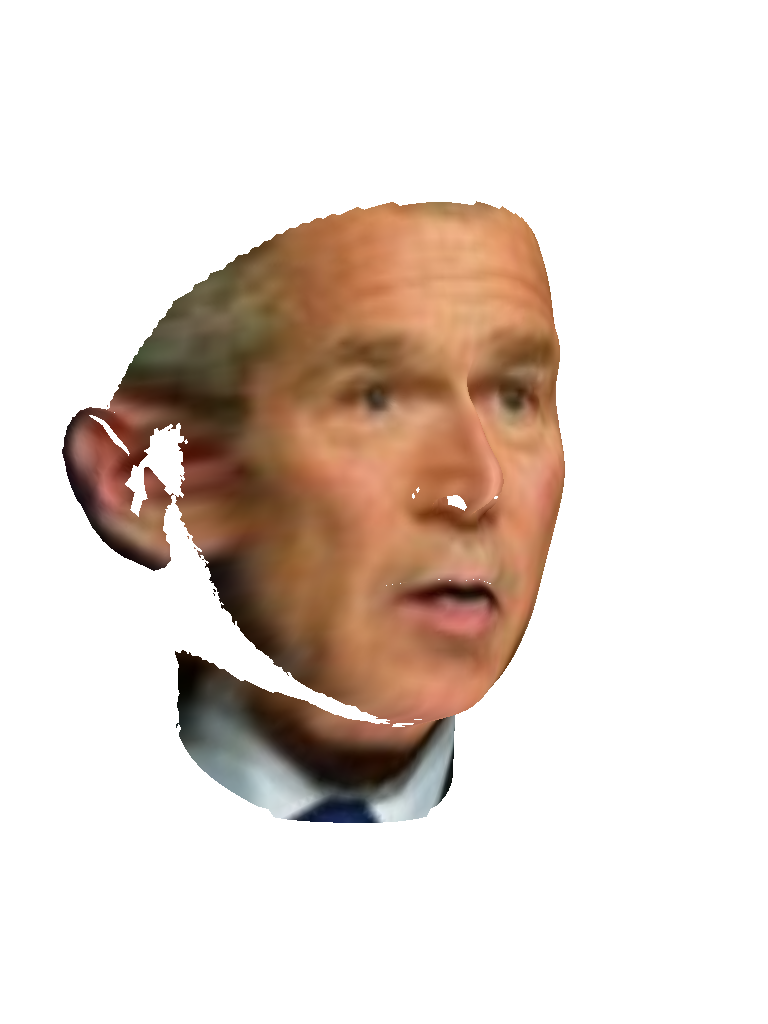}\\
\includegraphics[height=2.4cm, trim=60px 46px 67px 40px, clip=true]{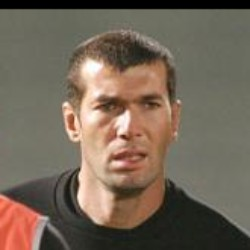}&
\includegraphics[height=2.4cm, trim=50px 210px 30px 130px, clip=true]{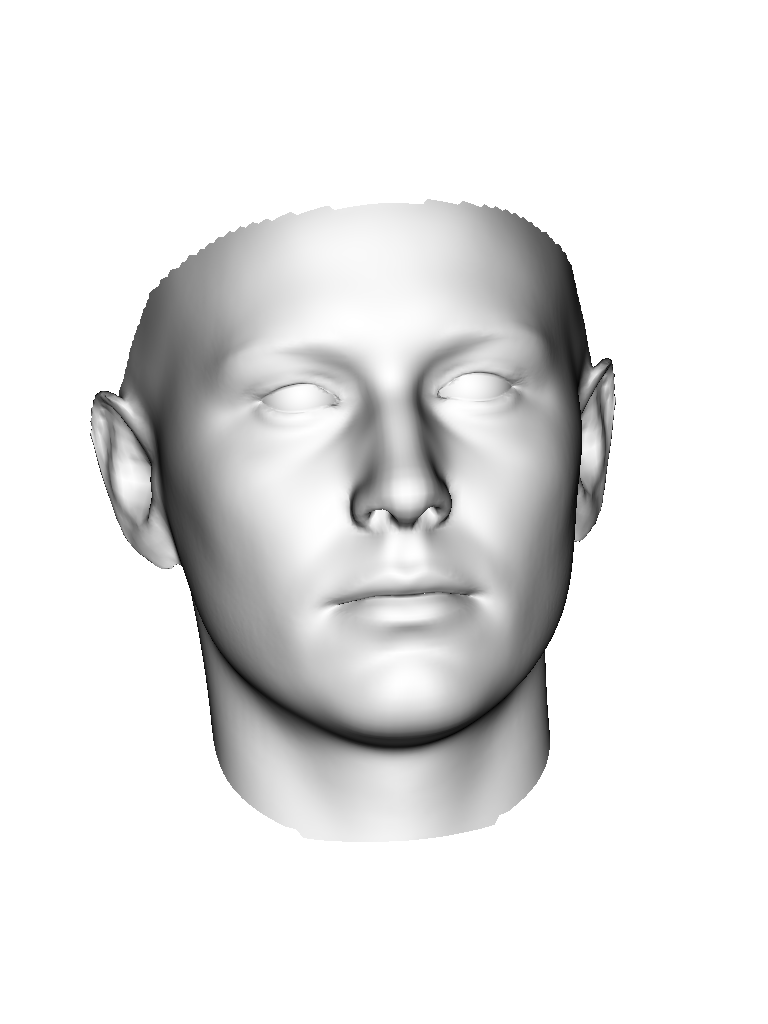}&
\includegraphics[height=2.4cm, trim=185px 140px 34px 167px, clip=true]{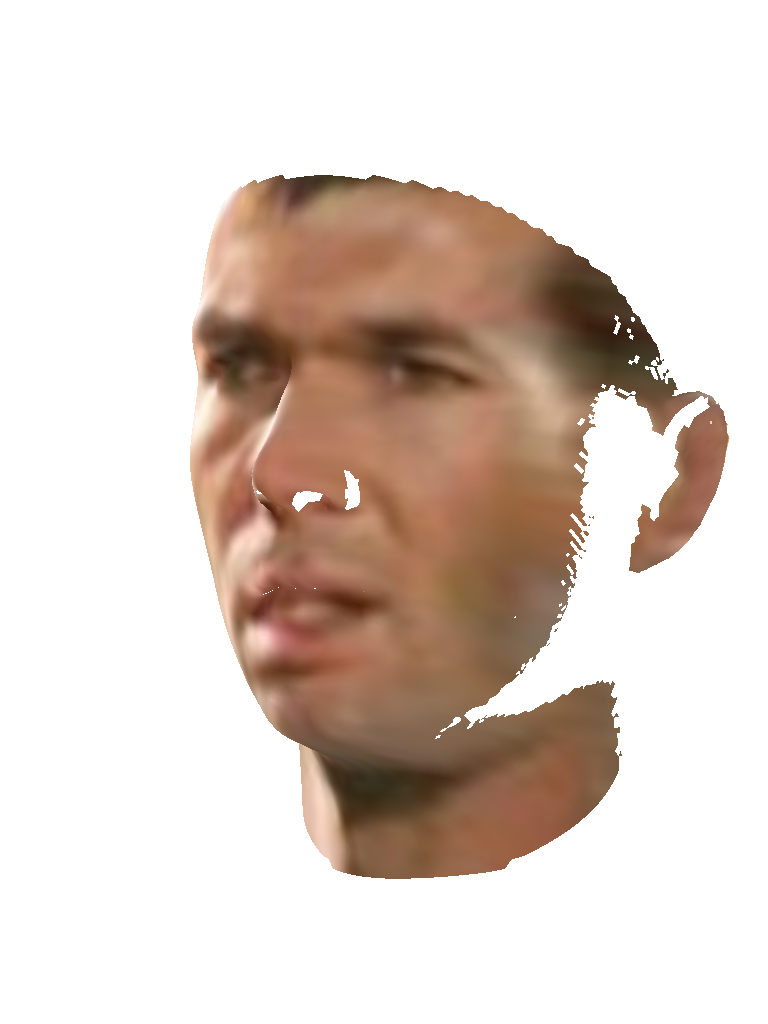}&
\includegraphics[height=2.4cm, trim=158px 140px 54px 167px, clip=true]{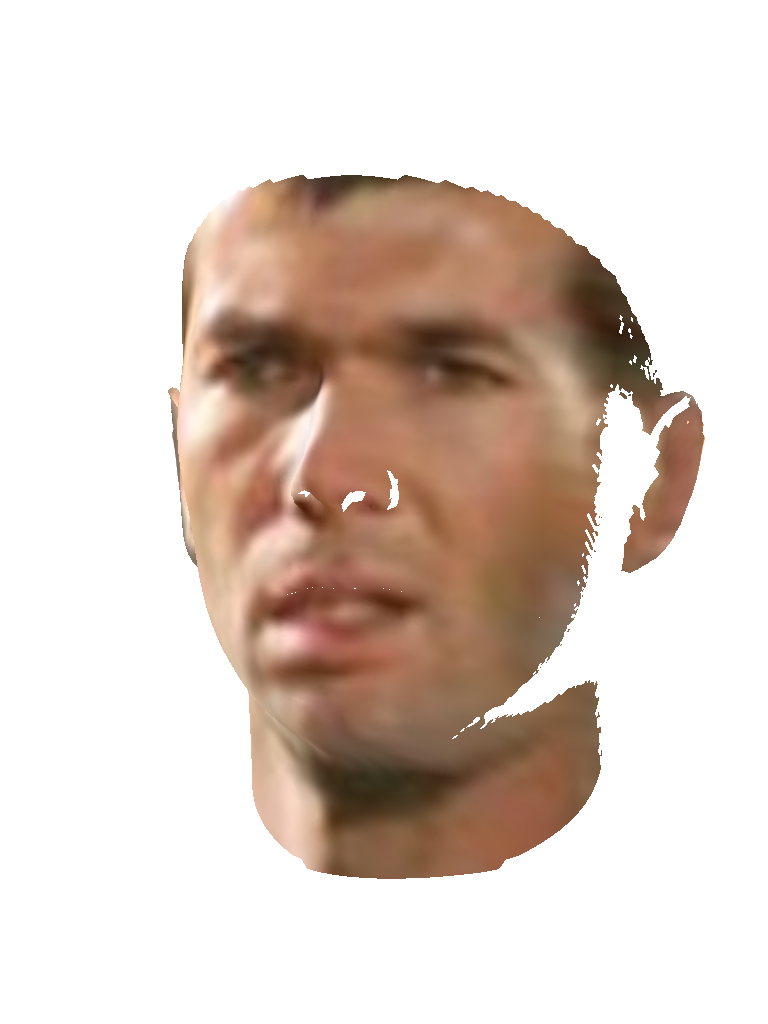}&
\includegraphics[height=2.4cm, trim=54px 140px 166px 167px, clip=true]{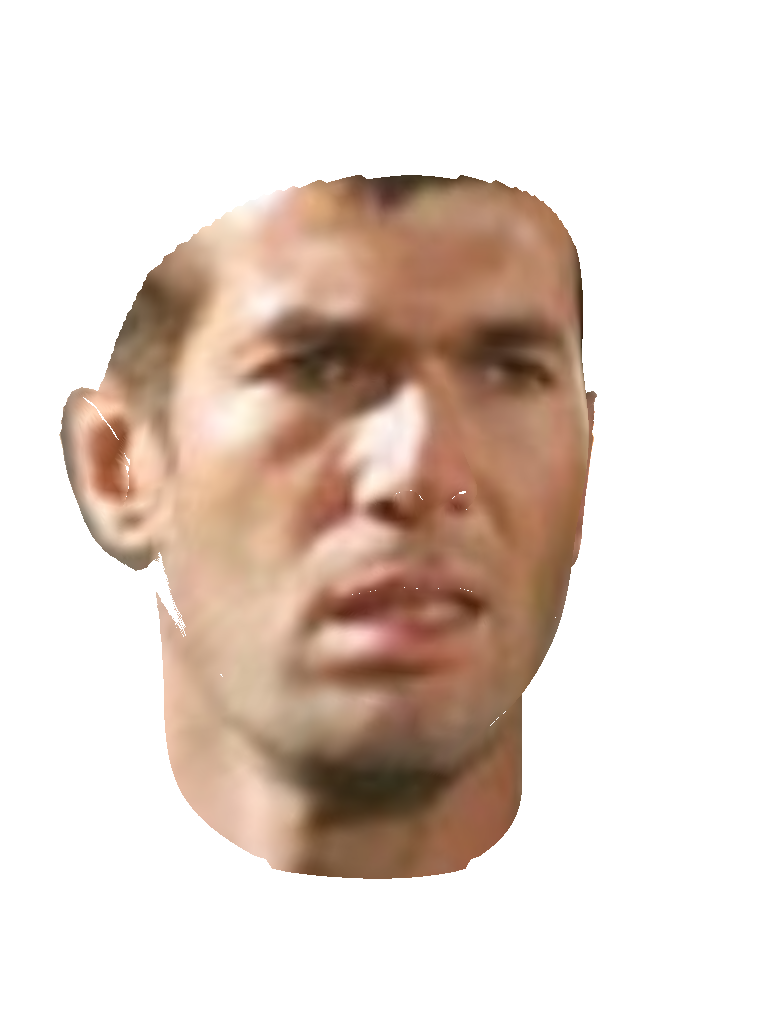}&
\includegraphics[height=2.4cm, trim=31px 140px 186px 167px, clip=true]{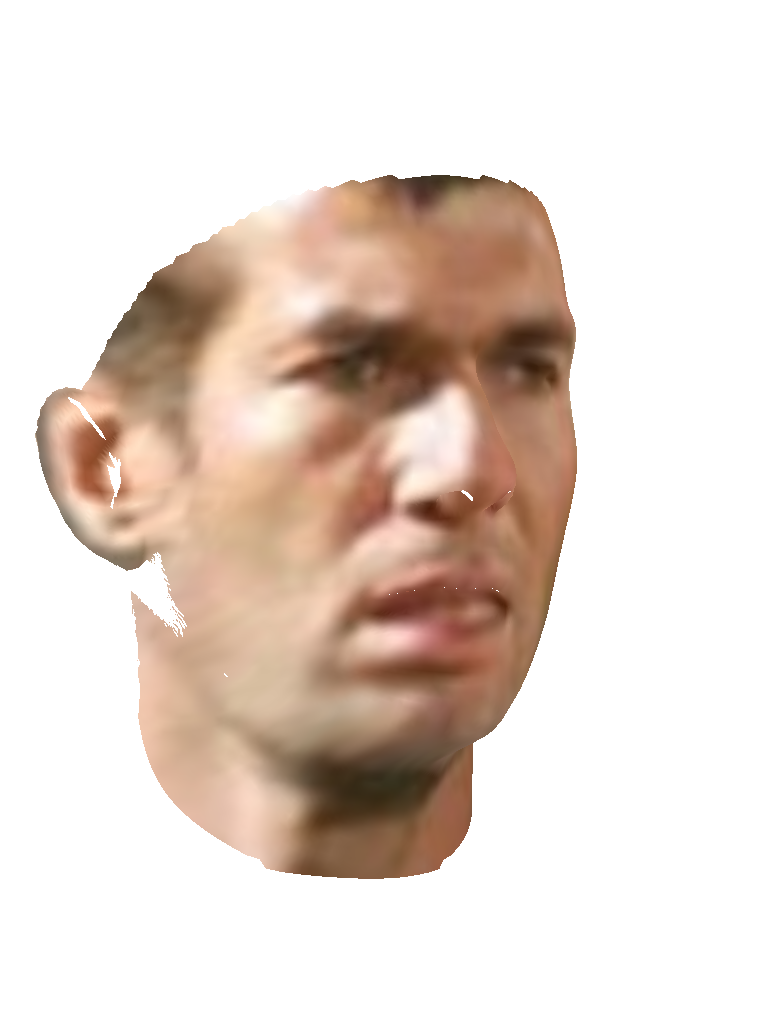}\\
\end{tabular}
\caption{Qualitative pose editing results}
\label{fig:LFW}
\end{figure}

In Figure \ref{fig:CMU} we show qualitative examples from the CMU PIE \cite{Sim:03} dataset. Here, we fit to images (first row) in a non-frontal pose using automatically detected landmarks \cite{zhu2012face} and show the reconstruction in the second row. We texture map the image onto the mesh, rotate to frontal pose (bottom row) and compare to an actual frontal view (third row). Finally, we show qualitative examples from the Labelled Faces in the Wild dataset \cite{LFWTech} in Figure \ref{fig:LFW}. Again, we texture map the image to the mesh and show a range of poses. These results show that our method is capable of robustly and fully automatically fitting to unconstrained images.

\setlength{\tabcolsep}{1pt}
\renewcommand{\arraystretch}{0.3}

\section{Conclusions}

We have presented a fully automatic algorithm for fitting a 3DMM to single images using hard edge correspondence and compared it to existing methods using soft correspondence.
In 3D-3D alignment, the soft correspondence of LM-ICP \cite{Fitzgibbon:03} is demonstrably more robust than hard ICP \cite{Besl:92}. However, in the context of 3D-2D nonrigid alignment, a soft edge cost function is neither continuous nor differentiable since contours appear, disappear, split and merge under parameter changes \cite{keller20073d}. This makes its optimisation challenging, unstable and highly dependent on careful choice of optimisation parameters. Although our proposed algorithm relies on potentially brittle hard correspondences, solving for shape and pose separately requires only solution of a linear problem and, together, optimisation of a multilinear problem. This makes iterated closest edge fitting very fast and it provides an initialisation that allows the subsequent nonlinear optimisation to converge to a better optimum. We believe that this explains the improved performance over edge fitting with soft correspondence.

There are many ways this work can be extended. First, we could explore other ways in which the notion of soft correspondence is formulated. For example, we could borrow from SoftPOSIT \cite{David:02} or Blind PnP \cite{Moreno:08} which both estimate pose with unknown 3D-2D correspondence. Second, we could incorporate any of the refinements to standard ICP \cite{Rusinkiewicz:01}. Third, we currently use only geometric information and do not fit texture. Finally, we would like to extend the method to video using a model that captures expression variation and incorporating temporal smoothness constraints.

\end{document}